\documentclass[10pt,journal,compsoc]{IEEEtran}
\ifCLASSOPTIONcompsoc
  \usepackage[nocompress]{cite}
\else
  \usepackage{cite}
\fi
\ifCLASSINFOpdf
\else
\fi
\usepackage{booktabs}

\usepackage{times}  
\usepackage{helvet} 
\usepackage{courier}  
\usepackage[hyphens]{url}  
\usepackage{graphicx} 
\usepackage{graphicx}  
\usepackage[utf8]{inputenc} 
\usepackage{url}            
\usepackage{booktabs}       
\usepackage{amsfonts}       
\usepackage{nicefrac}       
\usepackage{microtype}      
\usepackage{amsmath}
\usepackage{graphicx}
\usepackage{algorithm}
\usepackage{algorithmic}
\usepackage{amsmath}
\usepackage{amssymb}
\usepackage{multirow}
\usepackage{booktabs}

\newcommand{\R}{\mathbb{R}}

\DeclareMathOperator{\sut}{subject\ to\ \ \ \ }

\usepackage{amsmath}
\usepackage{tikz}
\usepackage{mathdots}
\usepackage{yhmath}
\usepackage{cancel}
\usepackage{color}
\usepackage{siunitx}
\usepackage{array}
\usepackage{multirow}
\usepackage{amssymb}
\usepackage{gensymb}
\usepackage{tabularx}
\usepackage{booktabs}
\usetikzlibrary{fadings}
\usepackage{subcaption}
\newcolumntype{H}{>{\setbox0=\hbox\bgroup}c<{\egroup}@{}}
\usepackage{adjustbox}

\makeatletter
\def\blfootnote{\xdef\@thefnmark{}\@footnotetext}
\makeatother

\begin{document}
%
\title{Deep Constraint-based Propagation in Graph Neural Networks}
%
%
%
%

\author{Matteo~Tiezzi,
       Giuseppe~Marra,
       Stefano~Melacci,~\IEEEmembership{Member,~IEEE,}
        and~Marco~Maggini
\IEEEcompsocitemizethanks{\IEEEcompsocthanksitem The authors are with the Deparment of Information Engineering and Mathematics, University of Siena, 53100 Siena, Italy.
\IEEEcompsocthanksitem Matteo Tiezzi is the corresponding author (mtiezzi@diism.unisi.it).}}
\IEEEtitleabstractindextext{%
\begin{abstract}

The  popularity of deep learning techniques renewed the interest in neural architectures able to process complex structures that can be represented using graphs, inspired by Graph Neural Networks (GNNs). { We focus our attention on the originally proposed GNN model of Scarselli et al. 2009,} which  encode{s} the state of the nodes of the graph by means of an iterative diffusion procedure that, during the learning stage, must be computed at every epoch, until the fixed point of a learnable state 
transition function is reached, propagating the information among the neighbouring nodes. 
We propose a novel approach to learning in GNNs, based on constrained
optimization in the Lagrangian framework. 
Learning both the transition function and the node states is the outcome of a joint process, in which the state convergence procedure is 
implicitly expressed by a constraint satisfaction mechanism, avoiding iterative epoch-wise procedures and the network unfolding.
Our computational structure searches for saddle points of the Lagrangian in the adjoint space composed 
of weights, nodes state variables and Lagrange multipliers. This process is further enhanced by multiple layers of constraints that accelerate the diffusion process.
An experimental analysis shows that the proposed approach compares favourably with popular models on several benchmarks.

\end{abstract}
}

\maketitle

\IEEEdisplaynontitleabstractindextext

%
\IEEEpeerreviewmaketitle

\IEEEraisesectionheading{\section{Introduction}\label{sec:intro}}

%
%
%
%
\blfootnote{Accepted for publication in IEEE Transactions on Pattern Analysis and Machine Intelligence (TPAMI) (DOI: 10.1109/TPAMI.2021.3073504).
}

\IEEEPARstart{N}{owadays} several real-world problems can be efficiently approached with Neural Networks, ranging from Computer Vision to Natural Language, from Bionformatics to Robotics. In the last decade, following the success of Deep Learning, neural models gained further popularly both in the scientific community and in the industry. Artificial Neural Networks are very flexible models with proved approximation capabilities, and their original processing and learning schemes have been extended in order to deal with structured inputs. While the original feed-forward model is able to process vectors of features as inputs, different architectures have been proposed to process sequences (Recurrent Neural Networks \cite{Williams1989ALA,hochreiter1997long}), groups of pixels (Convolutional Neural Networks \cite{LeCun:1998:CNI:303568.303704}), directed acyclic graphs (Recursive Neural Networks \cite{Goller96,Frasconi:1998:GFA:2325763.2326281}), and general graph structures (Graph Neural Networks \cite{DBLP:journals/tnn/ScarselliGTHM09}). These models generally share the same learning mechanism based on error BackPropagation (BP) through the network architecture, that allows the computation of the gradient of the cost function with respect to the connection weights. When processing structured data the original BP schema is straightforwardly extended by the process of {\em unfolding}, that generates a network topology based on the current input structure by replicating a base neural network module (e.g. BP Through Time, BP Through Structure).
However, recently, some authors \cite{carreira2014distributed,taylor2016training} proposed a different approach to learning neural networks, where neural computations are expressed as constraints and the optimization is framed into the Lagrangian framework. These algorithms are naturally local and they can be used to learn any computational structure, both acyclical or cyclical. The main drawback of these methods is that they are quite memory inefficient; in particular, they need to keep extra-variables for each hidden neuron and for each example. This makes them inapplicable to large problems where BP is still the only viable option. 

Graph Neural Networks (GNNs) are neural models that are able to process input data represented as graphs.
They exploit neural networks to learn task-dependent representations of the nodes of a graph, taking into account both the information that is locally attached to each node and the whole graph topology. { The seminal work by Scarselli et. al \cite{DBLP:journals/tnn/ScarselliGTHM09}, which henceforth will be denoted with GNN*, is} based on a pair of functions whose parameters (weights) are learned from data, that are the {\textit{state transition function} and the \textit{output function}}. {In a recent survey on Graph Neural Networks \cite{DBLP:journals/tnn/WuPCLZY21}, such model is classified as Recurrent GNN (RecGNN), due to the specific properties of their learning process}.  
{ In particular, the learning process of GNN*} requires, for each training epoch, an iterative diffusion mechanism that is repeated until it converges to a stable fixed point{ \cite{DBLP:journals/tnn/ScarselliGTHM09}}, requiring a multi-stage optimization procedure that is computationally expensive and less practical than models based on gradient-based optimization. The iterative process can be early stopped to speed-up the computation, but this ends up in limiting the quality of the outcome of the local encoding { learned by the GNN*}, virtually reducing the depth of the diffusion along the graph of the information carried by each node. In this paper, we propose a new learning mechanism for GNNs that is based on a Lagrangian formulation in which the relationship between each node and its neighborhood is represented by a set of constraints. Finding node state representations that fulfill the constraints is a simple way to rethink the computation of the fixed point of the aforementioned diffusion process. 
In particular, we can borrow tools from constrained optimization in Neural Networks \cite{Platt} to devise a learning scheme that is fully based on BP and where the  state representations and the weights of the state transition and output functions are jointly optimized without the need of applying any separate iterative procedures at every training epoch. 

Differently from the aforementioned Lagrangian-based training procedures \cite{carreira2014distributed,taylor2016training}, both the state transition function and the output function are classic BP-trainable models, that are shared by all the training examples, whereas the only additional variables of the learning problem are associated to the nodes of the graphs. This allows us to find a good trade off between the flexibility introduced by the Lagrangian-based formulation of the graph diffusion and the addition of new variables. We further extend this idea, originally presented in \cite{DBLP:conf/ecai/TiezziMMMG20},  computing multiple representations of each node by means of a pipeline of constraints that very much resembles a multi-layer computational scheme. In particular, the evolving representation of each node is treated as new information attached to the node itself. Another state transition function is introduced, that has the use of such new information, while constraints enforce a parallel diffusion process that leads to the development of another representation of the node. This procedure can be replicated multiple times, thus simulating a deep constraining scheme that augments the representation capabilities of the GNN. Experimental results on several popular benchmarks emphasize the quality of the proposed model, that compares favourably with the original GNN model and more recent variants.

The paper is organized as follows. Section \ref{sec:related} reviews the recent developments in the field of Neural Networks  for graphs and in Lagrangian approaches. Section \ref{sec:gnn} introduces the basics of the GNN model, whereas in Section \ref{sec:Lgnn} the proposed Lagrangian formulation of GNNs is described. Section \ref{sec:experiments} reports an experimental evaluation. Finally, conclusions are drawn in Section \ref{sec:concl}.

\section{Related Work}
\label{sec:related}

Despite the large ubiquity of data collected in Euclidean domains in which each sample is a fixed-length vector, a large number of application domains require to handle data that are characterized by an underlying structure that lays on a non-Euclidean domain, i.e. graphs \cite{DBLP:journals/spm/BronsteinBLSV17} and manifolds \cite{melacci2011primallapsvm}.
Early machine learning approaches for  structured data were designed for directed acyclic graphs \cite{Sperduti:1997:SNN:2325755.2326105, Frasconi:1998:GFA:2325763.2326281}, while a more generic framework ({GNN*}) was introduced in \cite{DBLP:journals/tnn/ScarselliGTHM09}. {GNN*} is able to deal with directed, undirected and cyclic graphs. The core idea behind {GNN*} is based on an iterative scheme of information diffusion among neighboring nodes, involving a propagation process aimed at reaching an equilibrium of the node states that represents a local encoding of the graph for a given task. Estimating such encoding is a computationally expensive process being based on the iterative computation of the fixed point of the state transition function. Some methods were aimed at simplifying this step, such as the scheme proposed in \cite{DBLP:journals/corr/LiTBZ15} that exploits gated recurrent units. { 
Steady State Embedding (SSE)  \cite{DBLP:conf/icml/DaiKDSS18} proved that algorithms on graphs can be effectively learned exploiting a constrained fixed-point formulation, solved via a \textit{policy iteration} algorithm in the context of Reinforcement Learning, and resorting to ad-hoc moving average updates to improve the transition and output functions.
In recent literature \cite{DBLP:journals/tnn/WuPCLZY21, DBLP:journals/nn/BacciuEMP20}, these pioneering methods are classified as RecGNNs, due to the iterative application of aggregation functions sharing the same set of  parameters. 

More recent trends in the field of Graph Representation learning leverage  models  composed of a limited number of layers parametrized by different weights in each layer. Several works highlighted the fact that such  models are Turing Universal if their capacity, defined as the product of the dimension of the state vector representation by the network depth (number of layers), is large enough {\cite{DBLP:conf/iclr/Loukas20}}.
However, stacking many layers of aggregation has the tendency of loosing node features information \cite{DBLP:conf/aaai/LiHW18}, an issue that can be overcomed with some tricks involving residual connections or particular node aggregation functions \cite{DBLP:journals/corr/abs-2003-00982}.
Such models are generally referred to as Convolutional GNNs (ConvGNNs)\cite{DBLP:journals/tnn/WuPCLZY21}, and they }
differ in the choice of the neighborhood aggregation method and the graph level pooling scheme, { in such a way that they can be categorized into \textit{spectral approaches} and \textit{spatial approaches}}. \textit{Spectral approaches} exploit particular embeddings of the graph { and spectral convolutions} \cite{DBLP:journals/corr/BrunaZSL13}. 
Simplified instances are based on smooth reparametrization \cite{DBLP:journals/corr/HenaffBL15} or approximation of the spectral filters by a Chebyshev expansion \cite{DBLP:conf/nips/DefferrardBV16}. { The filters of} Graph Convolutional Networks (GCNs) \cite{DBLP:conf/iclr/KipfW17} are restricted to operate in a 1-hop neighborhood of each node. 
{
\textit{Spatial methods} comprehend models such as  DCNNs \cite{DBLP:conf/nips/AtwoodT16},  GraphSAGE \cite{hamilton2017inductive}, PATCHY-SAN \cite{DBLP:conf/icml/NiepertAK16,  DBLP:conf/nips/DuvenaudMABHAA15}, DGCNN \cite{DBLP:conf/aaai/ZhangCNC18}, which directly leverage the graph topology, without the need of an intermediate representation. AWE \cite{ivanov2018anonymous} exploits kernel-based methods with a learning-based approach to learn graph embeddings.
Going beyond the aforementioned categories, there exist several other models inspired by the generalization of the convolution operation on graphs (see \cite{DBLP:journals/corr/abs-2005-03675} for a complete taxonomy), characterized by a finite number of message exchanges among neighbors. In \cite{DBLP:journals/tsp/GamaMLR19},  the \textit{selection} and \textit{aggregation} GNNs leverage linear shift invariant graph filters and temporally structured signals, respectively,  to create convolutional models capable to handle graph signals. In \cite{DBLP:journals/tsp/RuizGMR20}, convolutional filters able to that take into consideration the structure of the graph are obtained, leveraging trainable nonlinear activation functions and using graph median filters and graph max filters. These works fully take advantage of the representational power given by the multi-layered  parametrized aggregation process.

In an effort to unify the previously presented models (and others\cite{DBLP:journals/corr/abs-2003-00982}) under a common framework, Gilmer et al.  \cite{gilmer2017neural, gilmer2020message} introduced  Message Passing Neural Networks (MPNNs), leveraging the general concept of message exchange. Such framework comprises \cite{DBLP:journals/corr/abs-2003-00982} both \textit{isotropic} models, in which every neighbor equally contributes  to the state computation, and the \textit{anisotropic} ones, where  the model learns directly from data to weight each node contribution  (GAT\cite{velivckovic2018graph}, MoNet\cite{ monti2017geometric} and others \cite{bresson2017residual}). 


A parallel line of research focuses on  the analysis of the expressive capabilities of GNNs and their aggregation functions. 
The seminal work by Xu et al. \cite{DBLP:journals/corr/abs-1810-00826} paved the road to studies on the expressive power of GNNs through the notion of graph isomorphism. The authors propose a model, the Graph Isomorphism Network (GIN), having the same representational power of the Weisfeiler-Leman (WL) Test  \cite{weisfeiler1968reduction}. The main intuition is based on the usage of an injective aggregation function, e.g., the sum, which is able to distinguish and devise elements belonging to a multiset. Conversely, other types of aggregation functions tend to capture the proportion/distribution of elements of a given type (mean function) or ignore multiplicities, reducing the multiset to a simple set (max function).
However, the WL-test itself is not a sufficient condition for two graphs being isomorphic, and, for this reason, several other works try to capture higher-order graph properties \cite{morris2019weisfeiler,maron2018invariant}.
Bearing in mind these findings, other approaches try to combine several aggregation functions in order to devise a powerful representational   scheme  \cite{corso2020principal}}.

Departing from the GNN-related literature, our work also follows the path traced by those approaches that exploit optimization in the Lagrangian framework to train Neural Networks. A Lagrangian formulation of learning was studied in the seminal work of Yann LeCun \cite{lecun1988theoretical}, which proposed  a theoretical framework for { BP}. More recently, Carreira and Wang \cite{carreira2014distributed} introduced the idea of training networks whose architecture is described by a constraint-based representation, implying an increase of the variables involved in the optimization. Their optimization scheme is based on quadratic penalties, aiming at finding an approximate solution of the problem that is then refined in a post-processing phase.
Differently, \cite{taylor2016training} exploits closed-form solutions, where most of the architectural constraints are softly enforced, and further additional variables are introduced to parametrize the neuron activations. These approaches introduce a large flexibility in the learning process, making it \textit{local}, in the sense that the computations of the gradients are not the outcome of a { BP}-like scheme over the whole network, but they only depend on groups of neighbouring neurons/weights. 
As a result, the computations of different layers can be carried out in parallel. Inspired by these ideas, other approaches \cite{noia} exploit a block-wise optimization schema on the neural network architecture, { or they fully rely on local models based on the Lagrangian framework \cite{marra2020local}.}

In the proposed approach, we follow a novel mixed strategy that benefits from the simplicity of the Lagrangian-based approaches to overcome the aforementioned limitations of the state diffusion in { GNN*}. In particular, the majority of the computations still rely on { BP} while constraints are exploited only to define the diffusion mechanism. In order to intermix the { BP}-based modules with the constraints, we borrow tools for hard constraint optimization that were proposed in the context of Neural Networks \cite{Platt}. This allows us to carry out both the optimization of the neural functions and the diffusion process at the same time, instead of alternating them into two distinct phases (as in \cite{DBLP:journals/tnn/ScarselliGTHM09}), with a theoretical framework supporting the formulation (Lagrangian optimization).

\section{Graph Neural Networks}
\label{sec:gnn}

The term Graph Neural Network (GNN) refers to a general computational model that exploits the inference and learning schemes of neural networks to process non-Euclidean data, i.e., data organized as graphs.
Given an input graph $G=(V,E)$, where $V$ is a finite set of {\em nodes} and $E \subseteq V \times V$ collects the {\em arcs}, GNNs apply a two-phase computation on $G$. In the \textit{encoding} (or {\em aggregation}) phase the model computes a state vector for each node in $V$ by 
combining the states of neighboring nodes (i.e., nodes $u, v \in V$ that are connected by an arc $(u,v) \in E$). In the second phase, usually referred to as \textit{output} (or {\em readout}), the latent representations encoded by the states stored in each node are exploited to compute the model output. GNNs can implement either a \textit{node-focused} function, where an output is produced for each node of the input, or a \textit{graph-focused} function, where the representations of all the nodes are aggregated to yield a single output for the whole  graph.



{ 

Following the common framework of MPNNs \cite{gilmer2017neural,gilmer2020message},
let $x_v\in \R^s$ be the state (encoding) of node $v$, that is a vector with $s$ components. Each node can be eventually paired with a feature vector that represents node-related information that might be available in the considered task (often referred to as the {\em node label}), indicated with $l_v \in \R^m$.
Similarly, $l_{(v,u)} \in R^{d}$ is the eventually available feature vector attached to arc $(v,u) \in E$ (the {\em arc label}). Finally, $\mbox{pa}[v] = \{u \in V: (u,v) \in E \}$ is the set of the {\em parents} of node $v$ in $G$,  $\mbox{ch}[v] = \{u \in V: (v,u) \in E \}$ are the {\em children} of $v$ in $G$, $\mbox{ne}[v]=\mbox{pa}[v] \cup\mbox{ch}[v]$ are the {\em neighbors} of the node $v$ in $G$.
In MPNNs, the node state $x_v$ is updated by means of an iterative process, with iteration index $t$, that involves a message passing scheme among neighboring nodes $\mbox{ne}[v]$,  followed by an aggregation of the exchanged information. Formally,
\begin{eqnarray}
    \label{m1} x_{(v,u)}^{(t-1)} &=& \texttt{MSG}^{(t)}\left(x_v^{(t-1)}, x_u^{(t-1)}, l_{(v,u)}\right) \\
    \label{m2} x_v^{(t)} &=& \texttt{AGG}^{(t)}\left(x_v^{(t-1)}, \sum_{u \in ne[v]}  x_{(v,u)}^{(t-1)},  l_v\right) 
\end{eqnarray} 
where  $x_{(v,u)}^{(t)}$  is an explicit edge representation, i.e., the exchanged message, computed by a learnable map $\texttt{MSG}^t(\cdot)$. Function $\texttt{AGG}^t(\cdot)$ aggregates the messages from edges connecting the neighboring nodes, exploiting also local node information (state and features). 
The functions $\texttt{MSG}^{(t)}(\cdot)$ and $\texttt{AGG}^{(t)}(\cdot)$ are typically implemented via Multi-Layer Perceptrons (MLPs), whose  parameters (weights) are learned from data, and shared among nodes in order to gain   generalization capability and save memory consumption. The representation of each node right after the last time step represents the node-level output of the GNN.

The two functions of Eq.~(\ref{m1}) and Eq.~(\ref{m2}) might or might not be shared among different message passing iterations. In the former case the superscript $t$ in the function names is dropped, as in the original GNN* \cite{DBLP:journals/tnn/ScarselliGTHM09} (Section~\ref{sec:newg}). This setting is mostly related to the basic instance of what we propose in this paper (Section~\ref{sec:Lgnn}), that we will also extend to a stacked architecture composed of multiple layers (Section~\ref{sec:layered}). Differently, when there is no function sharing among the time steps, $\texttt{AGG}^{(a)}(\cdot)$ and $\texttt{AGG}^{(b)}(\cdot)$  are two distinct functions exploited at time steps $t=a$ and $t=b$, respectively (the same holds for $\texttt{MSG}$). In this case, the iterative message passing computations of Eq.~(\ref{m1}) and Eq.~(\ref{m2}) can be described as the outcome of a multi-layer model, in which, for example, the output of $\texttt{AGG}^{(t)}$ is provided as input to $\texttt{MSG}^{(t+1)}$. Hence an $\ell$-step message passing scheme can be seen as an $\ell$-layered model. This choice entails both a higher representational capability and increased space and computational complexities. However, when considering a fixed number of layers $\ell$, the model is able to directly propagate a message up-to an $\ell$-hop distance, as in ConvGNNs \cite{DBLP:journals/tnn/WuPCLZY21}.
We also mention that, in several modern models, the node state vector is usually initialized with the initial features, $x_v = l_v$, that is not necessarily the case of the original GNN* model \cite{DBLP:journals/tnn/ScarselliGTHM09}.}


{\subsection{GNN*}
\label{sec:newg}}


{ In the original GNN* model \cite{DBLP:journals/tnn/ScarselliGTHM09},} the state of node $v$ is the outcome of an iterative procedure in which a \textit{state transition function} $f_a$ processes information from the neighborhood of $v$ and from $v$ itself in order to compute a vector embedding. { Such function is implemented by a neural model with learnable parameters $\theta_{f_a}$}. Formally,
\begin{align}
        x_v^{(t)} &= f_a\left (x_{{\mbox{\scriptsize ne}}[v]}^{(t-1)}, l_{{\mbox{\scriptsize ne}}[v]}, l_{(v,{\mbox{\scriptsize ch}}[v])},l_{({\mbox{\scriptsize pa}}[v],v)}, x_{v}^{(t-1)}, l_{v}\ | \ \theta_{f_a} \right),
        \label{eq:aggregation}
\end{align}
where the three subscripts $\mbox{pa}[v],\mbox{ch}[v],\mbox{ne}[v]$, with an abuse of notation and for the sake of simplicity, provide a particular meaning to the symbol to which they are attached. In detail, $x_{{\mbox{\scriptsize ne}}[v]}$ represents the \textit{set} of the states of the nodes that are neighbours to node $v$, i.e., $\{x_u: u \in {\mbox{ne}}[v]\}$, and the same rationale is used in $l_{(v,{\mbox{\scriptsize ch}}[v])}$, and $l_{({\mbox{\scriptsize pa}}[v],v)}$.
It should be noted that the { function  $f_a$} may depend on a variable number of inputs, given that the nodes $v \in V$ may have different degrees $\mbox{de}[v]=|\mbox{ne}[v]|$. 
{In GNN* \cite{DBLP:journals/tnn/ScarselliGTHM09}, $f_a$ is computed as shown in the first row of Table~\ref{tab:gnn}. It is the outcome of summing-up the contributes of explicit edge representations obtained by jointly projecting the node-level information and the edge-level information. Such projection is computed by means of $h(\cdot|\theta_{h})$, an MLP that represents the learnable part of $f_a$. }
Existing implementations are usually invariant with respect to permutations of the nodes in $\mbox{ne}[v]$, unless some predefined ordering is given for the neighbors of each node.
\begin{table*}
    \centering
     
    \begin{tabular}{llc}
        \toprule
        Method: Function & Reference & Implementation of $f_a$\\
         \midrule
         {GNN*}: Sum & Scarselli et al. \cite{DBLP:journals/tnn/ScarselliGTHM09}   &  $\sum_{u \in \mbox{\tiny ne}[v]} h(x_u{^{(t-1)}},l_u,l_{(v,u)},l_{(u,v)},x_v{^{(t-1)}},l_v|\theta_h)$\\
         GIN: Sum  & Xu et al. \cite{DBLP:journals/corr/abs-1810-00826} & $h{^{(t)}} \big((1 + \epsilon)  x_v{^{(t-1)}} + \sum_{u \in \mbox{\tiny ne}[v]} x_u{^{(t-1)}}|\theta_h$ \big) \\
         
         GCN: Mean & Kipf and Welling \cite{DBLP:conf/iclr/KipfW17} & $\sigma \bigg( h_{W_0}^{(t)}  (x_v{^{(t-1)}|\theta_{W_0}}) + \sum_{u \in \mbox{\tiny ne}[v]} c_{u,v} h_{W_1}^{(t)} (x_u{^{(t-1)}|\theta_{W_1}})\bigg)$\\
         
         GraphSAGE: Max & Hamilton et al. \cite{hamilton2017inductive} & $\sigma \bigg( h^{(t)} \big( x_v^{(t-1)},  \max_{u \in \mbox{\tiny ne}[v]} x_u^{(t-1)}|\theta_{h}\big)\bigg)$ \\
         
         { GAT: Anisotropy} & {Veličković et al. \cite{velivckovic2018graph}  } & { $\sigma \big(\sum_{u \in \mbox{\tiny ne}[v]} \alpha^{(t-1)}_{u,v} h^{(t-1)}(x_u^{(t-1)}|\theta_h)\big)$}\\
         
         { Selection GNN: Linear Shift} & { Gama et al. \cite{DBLP:journals/tsp/GamaMLR19}  } & { $\sigma \big( \sum_{g=1}^{F^{(t-1)} }  H^{(t)}_{fg} x^{(t-1)}_g$ \big)}\\
         { Median GNN: Median} & { Ruiz et al. \cite{DBLP:journals/tsp/RuizGMR20}  } & { $ \sum_{k=0}^{\kappa}\big( w_{tk}^{f} \texttt{med}(S^k, \sum_{g=1}^{F^{(t-1)} }  H^{(t)}_{fg} x^{(t-1)}_g$\big) }\\
         
         \bottomrule\\
    \end{tabular}
   \caption{Common implementations of the state transition function $f_a$. {In GNN*, t}he function $h()$ is implemented by a feedforward neural network with $s$ outputs, whose input is the concatenation of its arguments (for example, in the first case the input consists of a vector of $2s+2m+2d$ entries, with $l_{(u,v)} \in \R^d$ and  $l_u \in \R^m$). For the sake of clarity, some of these formulas are reported in a simplified way w.r.t. the original proposal. 
   { In \cite{DBLP:conf/iclr/KipfW17}, $c_{u,v}$ denotes a normalization constant depending on node degrees, while $\alpha^{(t-1)}_{u,v}$ in \cite{velivckovic2018graph} is a learned attention coefficient which introduces anisotropy in the neighbor aggregation.   In the multi-hop linear combination from \cite{DBLP:journals/tsp/GamaMLR19}, $H^{(t)}$ denotes a subsampled linear shift invariant graph filter, $F^{(t)}$ is the number of features in layer $t$. A median graph filter, parametrized by $w_{tk}^{f}$ is leveraged by the median GNN \cite{DBLP:journals/tsp/RuizGMR20} to define nonlinear activation functions substituting pointwise functions. $S$ denotes the graph shift operator attached to the input graph, and $\kappa$ are the filter taps. See the papers in the reference column for details.}
   }
    \label{tab:gnn}
\end{table*}
Once the state transition function has been iterated $T$ times, a {GNN*} computes an output function on each node or on an aggregated representation of all the nodes of the graph (depending on the requirements of the task at hand), i.e.,
\begin{subequations}
\begin{align}
  \label{eq:readout_node}  y_v \ &= f_r\left(x_v^{(T)}\ |\ \theta_{f_r}\right),\\
  \label{eq:readout_graph} y_G \ &= f_r\left(\{x_v^{(T)}, v \in V\}\ |\ \theta_{f_r}\right),
\end{align}
\end{subequations}
where $y_v$ is the output in the node-focused case, whereas $y_G$ is the output in the graph-focused case.  
The recursive application of the state transition function $f_a()$ on the graph nodes yields a diffusion mechanism, whose range depends on $T$. In fact, by stacking $t$ times the aggregation of 1-hop neighborhoods by $f_a()$, information of one node can be transferred to the nodes that are distant at most $t$-hops. {{ In GNN*} }, Eq. (\ref{eq:aggregation}) is run until convergence of the state representation, i.e. until $x_v^{(t)} \simeq x_v^{(t-1)}, \forall v \in V$. This scheme corresponds to the computation of the {\em fixed point} of the state transition function $f_a()$ on the input graph. In order to guarantee the convergence of this phase, the transition function is required to be a {\em contraction map}\cite{DBLP:journals/tnn/ScarselliGTHM09}.  A sufficient number of iterations is the key to achieve a useful encoding of the input graph for the task at hand and, hence, the choice is problem-specific. 

{In order to contextualize GNN* in the MPNN framework, we notice that Eq.~(\ref{eq:aggregation}) actually includes both the main operations described in Eq.~(\ref{m1}) and Eq.~(\ref{m2}). As a matter of fact, $f_a(\cdot)$ can be seen as an instance of the $\texttt{AGG}^{(t)}(\cdot)$ function of Eq.~(\ref{m1}), that internally includes an arc-level message passing scheme $h(\cdot)$, essentially corresponding to $\texttt{MSG}^{(t)}(\cdot)$ of Eq.~(\ref{m2}). Various other aggregation mechanisms have been proposed in related approaches, as shown in Table~\ref{tab:gnn}.}
{ In GNN*, all the message passing iterations share the same $f_a$, so that Eq.~(\ref{eq:aggregation}) introduces a clear recurrent computational scheme that is the main reason why GNN* belongs to the category of RecGNN models \cite{DBLP:journals/tnn/WuPCLZY21}.
The multiple iterations of the fixed point computation can be seen as the \textit{depth} of the GNN*.
Another important feature of GNN* that distinguishes them from more recent implementations is that the node states $x_v \in \R^s$ are zero-initialized and independent from the eventually available node features $l_v \in  \R^m$ (i.e., $x_v^{(0)}$ is not equivalent to $l_v$). While using a shared function among multiple iterations might be lacking in terms of representational power \cite{DBLP:journals/tsp/GamaMLR19, DBLP:journals/tsp/RuizGMR20}, $f_a$ receives as input the eventually available node-level information ($l_k$) and the one attached to the arcs ($l_{(i,j)}$). This allows GNN* to avoid oversmoothing or loosing the connection with the available information about the considered graph \cite{DBLP:conf/aaai/LiHW18}.}

Henceforth,  for compactness, we denote the state transition function, applied to a node  $v \in V$,  with:
\begin{equation}
f_{a,v} = f_a\left(x_{{\mbox{\scriptsize ne}}[v]}, l_{{\mbox{\scriptsize ne}}[v]}, l_{(v,{\mbox{\scriptsize ch}}[v])},l_{({\mbox{\scriptsize pa}}[v],v)}, x_{v}, l_{v}\ | \ \theta_{f_a}\right).
\label{definisco}
\end{equation}
Basically, the encoding phase, through the iteration of $f_a()$, finds a solution to the fixed point problem defined by the \textit{equality constraint}
\begin{equation}
    \forall v \in V,\quad  x_v = f_{a,v}.
    \label{eq:fixedpoint}
\end{equation}
When this condition is met, the states virtually encode the information contained in the whole graph. 
{ The diffusion mechanism of GNN*, given the fact that convergence must be reached for every learning epoch, represents a computational burden for the model.}

\section{Constraint-based Propagation}
\label{sec:Lgnn}
When considering the learning stage of the original {GNN*} training algorithm, the computation of the fixed point (or of an approximation of it) is required at \textit{each epoch} of the learning procedure. The gradient computation requires to take into account such relaxation procedure, by a { BP} schema that involves the replicas of the state transition network exploited during the iterative fixed point computation ({\em unfolding}). 
This is due to the fact that the computation in {GNN*s} is driven by the input graph topology, that defines a set of constraints among the  state variables $x_v,\ v \in V$. In Section \ref{sec:gnn} we described how the fixed point computation aims at solving Eq.~(\ref{eq:fixedpoint}), that imposes an equality constraint between each node state and the way it is computed by the state transition function. 

Learning in Neural Networks can be also cast as a constrained optimization problem and solved in the Lagrangian framework \cite{carreira2014distributed}. The problem consists in the minimization of the classical data fitting loss (and eventually a regularization term) subject to a set of {\em architectural} constraints that describe the computation performed on the data (for example, in feed-forward networks we can constrain the variable associated to the output of a neuron to be equal to the value of the activation function computed on the weighted sum of the neuron inputs). The solution of the problem can be computed by finding the {\em saddle points} of the associated Lagrangian in the space defined by the  learnable parameters  and the Lagrange multipliers. 
In the specific case of {GNN*}, we can consider a Lagrangian formulation of the learning problem by adding free variables corresponding to the node states $x_v$, such that the fixed point is directly defined by the constraints themselves, as
\begin{align}
\forall v \in V,\quad  {\mathcal G}\left(x_v - f_{a,v}\right)=0,
\label{indiano}
\end{align}
where ${\mathcal G}(x)$ is a function characterized by $\mathcal{G}(0)=0$, such that the satisfaction of the constraints implies the solution of Eq.~(\ref{eq:fixedpoint}).
Apart from classical choices, like $\mathcal{G}(x) = x$ or   $\mathcal{G}(x) = x^2$, we can design different function shapes (see Section~\ref{sec:artificial}), with desired properties. For instance, a possible implementation is ${\mathcal G}(x)=\max(||x||_1-\epsilon,0)$, where $\epsilon \geq 0$ is a parameter that can be used to tolerate a small slack in the satisfaction of the constraint. The original formulation of the problem would require $\epsilon=0$, but by setting $\epsilon$ to a small positive value it is possible to obtain a better generalization and tolerance to noise.

In the following, for simplicity, we will refer to a node-focused task, such that for some (or all) nodes $v \in S \subseteq V$ of the input graph $G$, a target output $y_v$ is provided as a supervision\footnote{For the sake of simplicity we consider only the case when a single graph is provided for learning. The extension for more graphs is straightforward for node-focused tasks, since they can be considered as a single graph composed by the given graphs as disconnected components.}. If $L(f_r(x_v\ |\ \theta_{fr_r}),y_v)$ is the loss function used to measure the target fitting approximation for node $v \in S$, the formulation of the learning task is: 
\begin{align}
 \nonumber \min_{\theta_{f_a},\theta_{f_r}, X} \ \ \ \ & \sum_{v \in S} L(f_r(x_v\ |\ \theta_{f_r}), y_v)\\
\sut & \mathcal{G}\left(x_v - f_{a,v}\right) = 0, \quad \forall \ v \in V,
\label{eq:main_problem_1x}
\end{align}
where we already defined $\theta_{f_a}$ and $\theta_{f_r}$ as the weights of the MLPs implementing the state transition function and the output function, respectively, while $X = \{x_v: v \in V\}$ is the set of the newly introduced free state variables.

This problem statement implicitly includes the definition of the fixed point of the state transition function since for each solution in the feasible region the constraints are satisfied, and hence the learned $x_v$ are solutions of Eq.~(\ref{eq:fixedpoint}).
As introduced at the beginning of this Section, the constrained optimization problem of Eq.~(\ref{eq:main_problem_1x}) can be faced in the Lagrangian framework by including a Lagrange multiplier $\lambda_v$ for each constraint, while the Lagrangian function $\mathcal{L}(\theta_{f_a},\theta_{f_r}, X, \Lambda)$ is defined as:
\begin{multline}
\mathcal{L}(\theta_{f_a},\theta_{f_r}, X, \Lambda) = 
\sum_{v \in S} \left[ L(f_r(x_v\ |\ \theta_{f_r}), y_v) + \right. \\
\left. + \lambda_v \mathcal{G}\left(x_v - f_{a,v}\right)\right],
\label{eq:GNNlagrangian}
\end{multline}
where $\Lambda$ is the set of the $|V|$ Lagrangian multipliers. We can find solution of the learning problem by optimizing an unconstrained optimization index and searching for saddle points in the adjoint space 
$(\theta_{f_a},\theta_{f_r}, X, \Lambda)$. In detail, we aim at solving
\begin{align}
 \underset{\theta_{f_a},\theta_{f_r}, X}{\min}   \ \underset{\Lambda}{\max}\ \  \mathcal{L}(\theta_{f_a},\theta_{f_r}, X, \Lambda),
 \label{batmanx}
\end{align}
that can be approached with gradient descent-based optimization with respect to the variables $\theta_{f_a},\theta_{f_r}, X$, and gradient ascent with respect to the Lagrange multipliers $\Lambda$ (see \cite{Platt}).
Interestingly, the gradient can be computed locally to each node, given the node-related variables and those of the neighboring nodes. In fact, the derivatives of the Lagrangian with respect to the involved parameters are\footnote{When parameters are vectors, the reported gradients should be considered element-wise.}:
\begin{align}
\label{cazzo}\frac{\partial \mathcal{L}}{\partial x_v} & = L'f'_{r,v} + \lambda_v \mathcal{G}'_v (1 - f'_{a,v}) - \hskip-2mm \sum_{w:v \in \mbox{\scriptsize ne}[w]} \hskip-2mm \lambda_w \mathcal{G}'_w f'_{a,w} \\
\label{stocazzo}\frac{\partial \mathcal{L}}{\partial \theta_{f_a}} & = - \sum_{v \in S} \lambda_v \mathcal{G}'_v f'_{a,v} \\
\label{stograndissimocazzo}\frac{\partial \mathcal{L}}{\partial \theta_{f_r}} & = \sum_{v \in S} L'f'_{r,v} \\
\label{eq:gradient_lambda}
\frac{\partial \mathcal{L}}{\partial \lambda_v} & = \mathcal{G}_v 
\end{align}
where $f_{a,v}$ is defined in Eq. (\ref{definisco}), $f'_{a,v}$ is its  derivative\footnote{In Eq. (\ref{cazzo}) and Eq. (\ref{stocazzo}) such derivative is computed with respect to the same argument as in the partial derivative on the left side.}, $f_{r,v} = f_r(x_v\ |\ \theta_{f_r})$, $f'_{r,v}$ is its  derivative (with respect to $\theta_{f_r})$, $\mathcal{G}_v = \mathcal{G}\left(x_v -f_{a,v}\right)$ and $\mathcal{G}'_v$ is its first derivative, and, finally, $L'$ is the first derivative of $L$. Being $f_{a}$ and $f_{r}$ implemented by feedforward neural networks, their derivatives are obtained easily by applying a classic { BP} scheme, in order to optimize the Lagrangian function in the descent-ascent procedure, aiming at reaching a saddle point, following \cite{Platt}.
Even if the proposed formulation adds the free state variables $x_v$ and the Lagrange multipliers $\lambda_v$, $v \in V$, there is no significant increase in the memory requirements since the state variables also need to be memorized in the original formulation of {GNN*}, and there is just a single Lagrange multiplier for each node.

This novel approach in training {GNN*s} has several interesting properties.
The learning algorithm is based on a mixed strategy where \textit{(i.)} { BP} is used to efficiently update the weights of the neural networks that implement the state transition and output functions, and  \textit{(ii.)} the diffusion mechanism evolves gradually by enforcing the convergence of the state transition function to a fixed point by virtue of the constraints. This introduces a significant difference with respect to running an iterative procedure at each epoch and, only afterwards, applying the backward stage of { BP} to update the weights of $f_a$ and $f_r$, as done {in GNN*}. In the proposed scheme, { which we denote with the name Lagrangian Propagation GNNs (LP-GNNs),} both the neural network weights and the node state variables are simultaneously updated by gradient-based rules. The learning proceeds by jointly updating the function weights and by diffusing information among the nodes, through their state, up to a stationary condition where both the objective function is minimized and the state transition function has reached a fixed point of the diffusion process. This also introduces a strong simplification in the way the algorithm can be implemented in modern software libraries that commonly include automatic gradient computation.


\subsection{{Deep LP-GNNs}}
\label{sec:layered}
The {GNN*} computation of Eq.~(\ref{eq:aggregation}) may exploit a Multi-Layer Neural Network with any number of hidden layers to implement the state transition function $f_a$. By using more layers in this network, the model is able to learn more complex functions to diffuse the information on the graph, but the additional computation is completely local to each node. A different approach to yield a deep structure is to add layers to the state computation mechanism, in order to design a Layered {GNN*} \cite{Bianchini2018}. {Differently from Section~\ref{sec:gnn}, in which we emphasized the recurrence over index $t$ and its  interpretation in the framework of MPNNs, in this section we refer to physically stacking multiple instances of GNN*s to get Layered GNN*s}. Basically, a set of $K$ states $\left\{x_{v,k}, k=0,\ldots,K-1\right\}$ is computed for each node $v \in V$. The state of the first layer, $x_{v,0}$, is computed by Eq.~(\ref{eq:aggregation}) and it becomes a labeling of node $v$ that can be used to computed the state of the node at the following layer, i.e., $x_{v,1}$. More generally, at layer $k+1$ we have the use of the node label $l_v^{k+1}=x_{v,k}$. A different state transition function $f^k_a$ may be exploited at each layer. Formally, the computation is performed by the following schema for each layer $k>0$:
\begin{align}
        x_{v,k}^{(t)} &= f_a^k\left (x_{{\mbox{\scriptsize ne}}[v],k}^{(t-1)}, x_{v,k}^{(t-1)}, x_{v,k-1}\ | \ \theta_{f^k_a} \right),
        \label{eq:layered_aggregation}
\end{align}
whereas the states of the first layer $x_{v,0}$ are computed by Eq.~(\ref{eq:aggregation}). 
The model outputs are computed by applying the output network $f_r$ to the states $x_{v,K-1}$ available at the last layer. The proposed implementation is a simplification of the more general model that may process the neighboring node and arc labels $l_{{\mbox{\scriptsize ne}}[v]}, l_{(v,{\mbox{\scriptsize ch}}[v])},l_{({\mbox{\scriptsize pa}}[v],v)}$ (and also additional node labels to augment $l_{v}$) again at each layer\footnote{The model can be also extended  by considering an output function $f_r^k$ for each layer such that the output $y_{n,k-1}$ at layer $k-1$ is concatenated to $x_{v,k-1}$ for each node as input for the following layer. These intermediate outputs can be subject to the available supervisions \cite{Bianchini2018}.}. 

Following the original {GNN*} computation schema, the states need to be sequentially computed layer-by-layer applying the relaxation procedure to reach the fixed point: when moving to layer $k$ the states computed by the previous layer $k-1$ are considered constant inputs, as shown in Eq.~(\ref{eq:layered_aggregation}). As a result, Layered {GNN*} may be computationally demanding since they require to compute the fixed point for each layer in the forward phase, and to backpropagate the information through the resulting unfoldings in the backward phase.
In order to overcome these issues, we can exploit the locality in the computations of the proposed {LP-GNN} by considering the state variables $x_{v,k}$ at each node and layer as free variables. Once we introduce new layer-wise constraints with the same structure of Eq.~(\ref{indiano}), the proposed learning problem can be generalized to multiple layers as follows:
\begin{align}
 \nonumber \min_{\Theta_{f_a},\theta_{f_r}, X} &  \sum_{v \in S} L(f_r(x_{v,K-1}\ |\ \theta_{f_r}),y_v)\\
 \sut & \mathcal{G}(x_{v,k} - f^k_{a,v}) = 0, \quad \forall \ v \in V,\ \forall \ k \in [0,K-1]
\label{eq:main_problem_1}
\end{align}
where $\Theta_{f_a}=\left[\theta_{f^0_a},\ldots,\theta_{f^{K-1}_a}\right]$ collects the weights of the neural networks implementing the transition function of each layer, and $X$ collect the states $x_{v,k}$ for each node and layer. The notation $f^k_{a,v}$ is a straightforward extension of the one proposed in Eq. (\ref{definisco}), taking into account the schema defined by Eq.~(\ref{eq:layered_aggregation}), where  $f_{a,v}^k$, with $k>0$, is function of $x_{v}^{k-1}$. In this context,   $f^0_{a,v}$ corresponds to the original definition.
Our approach not only allows us to jointly optimize the weights of the networks and diffuse the information along the graph, but also to propagate the information through the layers, where, for each of them, a progressively more informed diffusion process is carried on and optimized. 

{
\subsection{Complexity and Expressiveness}
\label{sec:com_expr}

Common RecGNNs models \cite{DBLP:journals/tnn/WuPCLZY21}, to which GNN* belongs, exploit synchronous updates among all nodes and multiple iterations for the node state embedding, with a computational complexity for each parameter update of $\mathcal{O}(T(|V|+ |E|))$, where $T$ is the number of iterations, $|V|$ the number of nodes and $|E|$ the number of edges \cite{DBLP:conf/icml/DaiKDSS18}. By simultaneously carrying on the optimization of neural models and the diffusion process, the LP-GNN scheme proposed in this paper relies solely on 1-hop neighbors for each parameter update, hence showing a computational cost of $\mathcal{O}(|V|+ |E|)$. The potential addition of further layers of constraints causes a proportional increase of the complexity.  From the point of view of memory usage, the persistent state variable matrix requires $\mathcal{O}(|V|)$ space for each layer.

Many different GNN models have been recently analysed in terms of their ability to discriminate non-isomorphic graphs that the classical WL algorithm \cite{weisfeiler1968reduction} is able to discriminate.
The  GNN* model \cite{DBLP:journals/tnn/ScarselliGTHM09} has been shown to belong to the class of MPNNs, which, in turn, are showed to be at most as expressive as the 1-WL test \cite{DBLP:journals/corr/abs-1810-00826,morris2019weisfeiler}. The Lagrangian formulation of learning explored in this paper represents a different parametrization of the classical GNN*s, with a newly introduced optimization framework. However, at convergence, the functional model of LP-GNN is fully equivalent to the one of the original GNN* framework, thus it inherits the same expressiveness of GNN*.
As a result, we argue that the Lagrangian approach of this paper could be implemented exploiting the aggregation processes of other models (e.g., GCN \cite{DBLP:conf/iclr/KipfW17}, GIN \cite{DBLP:journals/corr/abs-1810-00826}), yielding different perspectives in term of expressiveness.


}

\section{Experiments}
\label{sec:experiments}

{}
We implemented the algorithm described in the previous sections using TensorFlow\footnote{\url{https://www.tensorflow.org}}. The implementation exploits the TensoFlow facilities to compute the gradients (Eq. (\ref{cazzo})-(\ref{eq:gradient_lambda})), while we updated the parameters of the problem of Eq. (\ref{batmanx}) using the Adam optimizer \cite{kingma2014adam}. For the comparison with the original GNN model, we
exploited the GNN Tensorflow implementation\footnote{The framework is available at \url{https://github.com/mtiezzi/gnn} and \url{https://github.com/mtiezzi/lpgnn}. The documentation is available at \url{http://sailab.diism.unisi.it/gnn/}}
introduced in \cite{rossi2018inductive}.
As it will be further investigated in Section \ref{sec:depth_vs_diffusion}, 
in our proposed algorithm, the diffusion process is turned itself into an optimization process that must be carried out both when learning and when making predictions.
As a matter of fact,
inference itself requires the diffusion of information through the graph, that, in our case, corresponds with satisfying the constraints of Eq.~(\ref{indiano}). For this reason, the testing phase requires a (short) optimization routine to be carried out, that simply looks for the satisfaction of Eq.~(\ref{indiano}) for test nodes, and it is implemented using the same code that is used to optimize Eq. (\ref{batmanx}), avoiding to update the state transition and output functions.

\subsection{Qualitative analysis}

\label{sec:qualitative}

When analyzing graphs, a very interesting analysis \cite{DBLP:conf/iclr/KipfW17} consists in having a look at how latent representations of nodes (states) evolve during the learning process. Indeed, when the dependence of the state transition function on the available node-attached features ($l_v^0$) is reduced or completely removed, the algorithm can only rely on the topology of the graph to perform the classification task at hand. In this setting, the states are continuous representations of topological features of the nodes in the graph and the LP-GNN model would implicitly learn a metric function in this continuous space. For this reason, we would expect that nodes belonging to the same class are placed close to each other in the embedding space.

\begin{figure}
    \centering
    \includegraphics[width=0.75\linewidth]{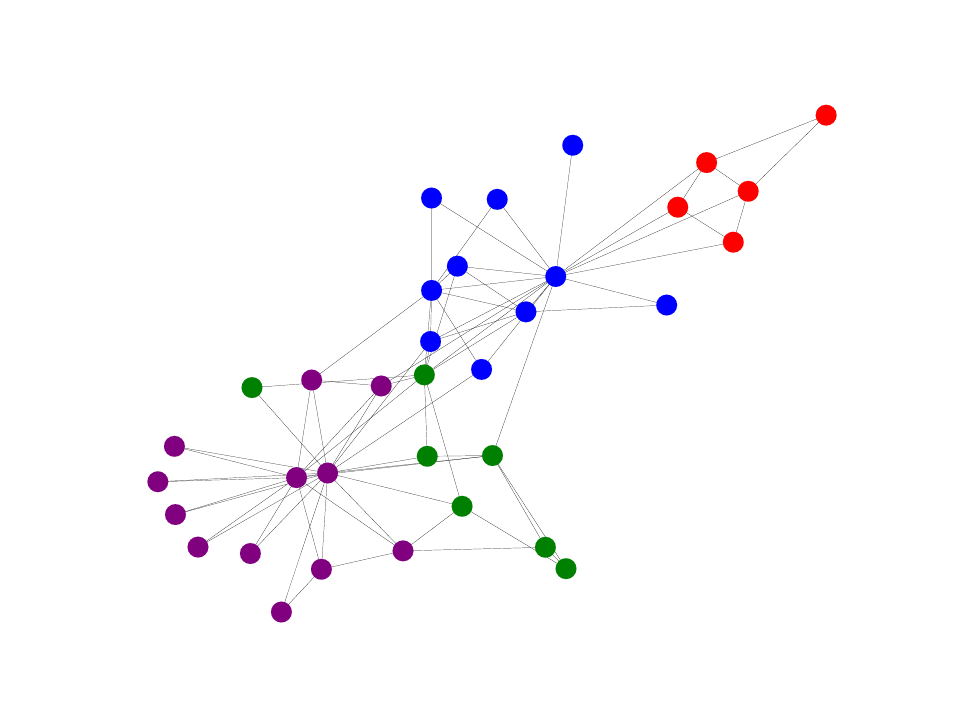}\\
    \vskip -6mm 
    \caption{{The karate club dataset.} This is a simple and well-known dataset exploited to perform a qualitative analysis of the behaviour of our model. Nodes have high intra-class connections and low inter-class connections. Each color is associated to a different class (4 classes). 
    }
    \label{fig:karate_data}
\end{figure}

In order to perform this evaluation, we exploit the simple and well-known Zachary Karate Club dataset \cite{zachary1977information}. The data was collected from the members of a university karate club by Wayne Zachary in 1977. Each node represents a member of the club, and each edge represents a tie between two members.  There are 34
nodes, connected by 154 (undirected and unweighted) edges. Every node is labeled by one out of
four classes, obtained via modularity-based clustering (see Figure \ref{fig:karate_data}).

\begin{figure*}
  \centering
  \medskip
  \begin{subfigure}[t]{.33\linewidth}
    \centering\includegraphics[width=\linewidth]{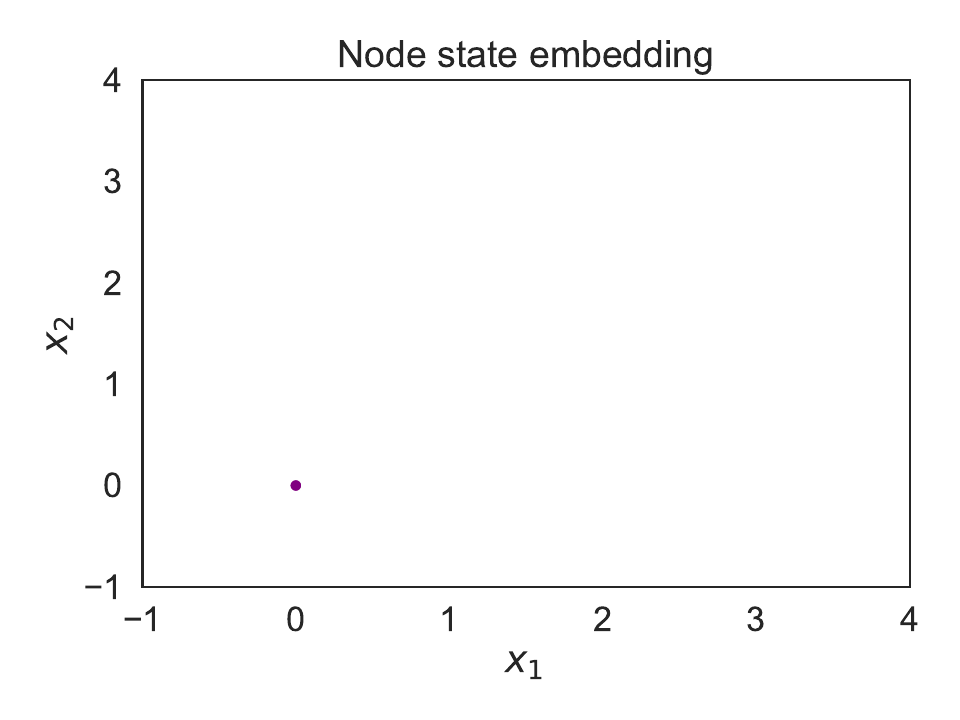}\\
    \vskip -4mm
    \caption{}
    \label{fig:karate:beg}
  \end{subfigure}
  \begin{subfigure}[t]{0.33\linewidth}
 \centering\includegraphics[width=\linewidth]{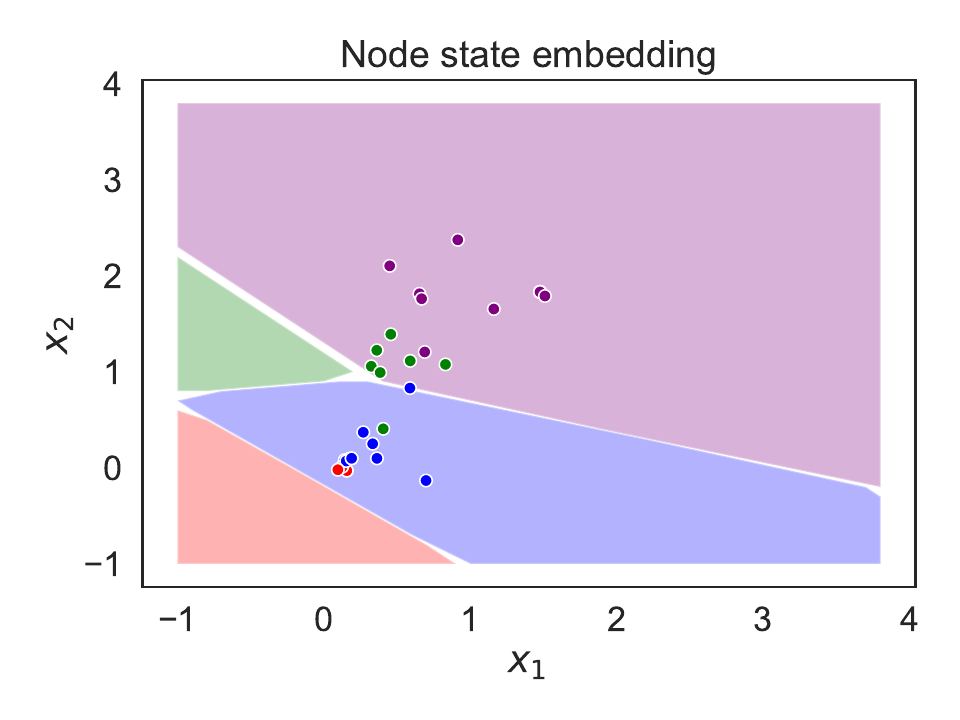}\\
 \vskip -4mm
    \caption{}
    \label{fig:karate:mid}
  \end{subfigure}
  \begin{subfigure}[t]{.33\linewidth}
 \centering\includegraphics[width=\linewidth]{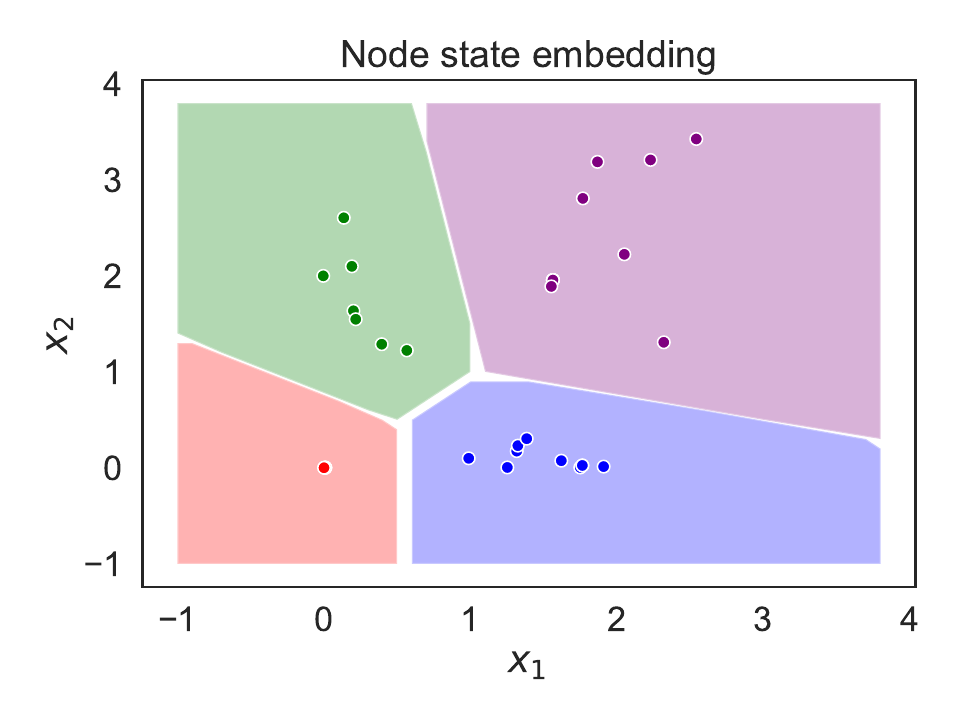}\\
 \vskip -4mm
    \caption{}
    \label{fig:karate:end}
  \end{subfigure}
  \vskip -1mm
  \caption{{Node state embeddings.} Evolution of the node state embeddings at different stages of the learning process: (\subref{fig:karate:beg})  beginning, (\subref{fig:karate:mid}) after 200 epochs and (\subref{fig:karate:end}) at convergence. We are not exploiting any node-attached features from the available data, so that the plotted node representations are the outcome of the diffusion process only, which is capable of mapping the topology of the graph into meaningful latent representations. Each node is represented with the color of the given corresponding class (ground truth), while the four background colors are the predictions of the output function learned by our model. The model learns node state embeddings that are linearly separated with respect to the four classes.}
  \label{fig:karate}
  
\end{figure*}

We trained a layered LP-GNN with three state layers ($K=3$). In order to visualize the node states and how they change over time, we set the state dimension of the last layer to 2 units and we used a shallow softmax regressor as output function, to force a linear separation among classes. Moreover, node-attached features of the given data were totally removed in order to force the algorithm to exploit only structural properties in the solution of the classification task.
Figure \ref{fig:karate} shows how the node states evolve over time, starting from an initialization composed of zero-only states (i.e. node states are initialized to a zero value) and then they move progressively toward four distinct areas of the 2D embedding space, one for each class. Since features of nodes were removed, the distinct areas of the space group nodes only with similar topological features.

\subsection{Artificial Tasks}
\label{sec:artificial}
We consider the two tasks of subgraph matching and clique localization. These tasks represent different challenges for the proposed model. { In this experiment, the LP-GNN must be able to leverage both the topology of the input graph and the available features $l_v$ attached to each node, that will be described in the following.} We first describe the main {properties} of the considered tasks, and, afterward, we describe the experimental results.

\begin{table}[th!]
 \caption{The considered variants of the  $\mathcal{G}$ function. By introducing $\epsilon$-insensitive constraint satisfaction, we can inject a controlled amount (i.e., $\epsilon$) of tolerance of the constraint satisfaction into the hard-optimization scheme (\textit{lin}: $\mathcal{G}(x)=x$;  \textit{lin-$\epsilon$}: $\mathcal{G}(x)=\max(x, \epsilon) - \max(-x, \epsilon)$; \textit{abs}: $\mathcal{G}(x)=|x|$;  \textit{abs-$\epsilon$}: $\mathcal{G}(x)=\max(|x|- \epsilon, 0)$; \textit{squared}: $\mathcal{G}(x)=x^2$).}
    \centering
    \begin{tabular}{c|c|c|c|c|c}
    \toprule

        \multirow{1}{*}{Function $\mathcal{G}(x)$}  & \textit{lin} & \textit{lin-$\epsilon$} & \textit{abs} & \textit{abs}-$\epsilon$ & \textit{squared}\\
         \toprule           
          Unilateral & $\times$ &$\times$ & $\checkmark$ & $\checkmark$ & $\checkmark$\\
         $\epsilon$-insensitive & $\times$ & $\checkmark$ & $\times$ & $\checkmark$ & $\times$\\
         \bottomrule
    \end{tabular}

      \label{tab:g_functions}
\end{table}

\paragraph{\textbf{Subgraph Matching}}

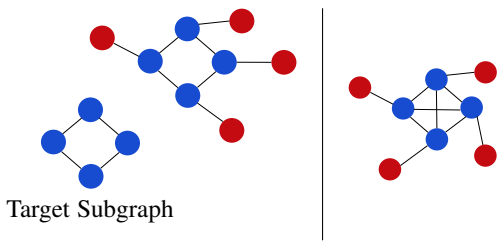
\begin{figure}
\centering

\tikzset{every picture/.style={line width=0.3pt}} 

\begin{tabular}{c@{\quad}|@{\quad}c}
\begin{tikzpicture}[x=0.2pt,y=0.2pt,yscale=-1,xscale=1]

\draw  [color={rgb, 255:red, 0; green, 0; blue, 0 }  ,draw opacity=0 ][fill={rgb, 255:red, 23; green, 76; blue, 208 }  ,fill opacity=1 ] (135,197.64) .. controls (135,184.73) and (145.59,174.27) .. (158.65,174.27) .. controls (171.71,174.27) and (182.3,184.73) .. (182.3,197.64) .. controls (182.3,210.54) and (171.71,221) .. (158.65,221) .. controls (145.59,221) and (135,210.54) .. (135,197.64) -- cycle ;
\draw  [color={rgb, 255:red, 0; green, 0; blue, 0 }  ,draw opacity=0 ][fill={rgb, 255:red, 23; green, 76; blue, 208 }  ,fill opacity=1 ] (205,260.64) .. controls (205,247.73) and (215.59,237.27) .. (228.65,237.27) .. controls (241.71,237.27) and (252.3,247.73) .. (252.3,260.64) .. controls (252.3,273.54) and (241.71,284) .. (228.65,284) .. controls (215.59,284) and (205,273.54) .. (205,260.64) -- cycle ;
\draw  [color={rgb, 255:red, 0; green, 0; blue, 0 }  ,draw opacity=0 ][fill={rgb, 255:red, 23; green, 76; blue, 208 }  ,fill opacity=1 ] (65,258.64) .. controls (65,245.73) and (75.59,235.27) .. (88.65,235.27) .. controls (101.71,235.27) and (112.3,245.73) .. (112.3,258.64) .. controls (112.3,271.54) and (101.71,282) .. (88.65,282) .. controls (75.59,282) and (65,271.54) .. (65,258.64) -- cycle ;
\draw  [color={rgb, 255:red, 0; green, 0; blue, 0 }  ,draw opacity=0 ][fill={rgb, 255:red, 23; green, 76; blue, 208 }  ,fill opacity=1 ] (136,323.64) .. controls (136,310.73) and (146.59,300.27) .. (159.65,300.27) .. controls (172.71,300.27) and (183.3,310.73) .. (183.3,323.64) .. controls (183.3,336.54) and (172.71,347) .. (159.65,347) .. controls (146.59,347) and (136,336.54) .. (136,323.64) -- cycle ;
\draw    (174.9,212.84) -- (211.9,244.84) ;

\draw    (103.9,276.84) -- (140.9,308.84) ;

\draw    (178,308.28) -- (213.05,278.3) ;

\draw    (106,243.28) -- (142.05,213.3) ;

\draw  [color={rgb, 255:red, 0; green, 0; blue, 0 }  ,draw opacity=0 ][fill={rgb, 255:red, 23; green, 76; blue, 208 }  ,fill opacity=1 ] (318,44.64) .. controls (318,31.73) and (328.59,21.27) .. (341.65,21.27) .. controls (354.71,21.27) and (365.3,31.73) .. (365.3,44.64) .. controls (365.3,57.54) and (354.71,68) .. (341.65,68) .. controls (328.59,68) and (318,57.54) .. (318,44.64) -- cycle ;
\draw  [color={rgb, 255:red, 0; green, 0; blue, 0 }  ,draw opacity=0 ][fill={rgb, 255:red, 23; green, 76; blue, 208 }  ,fill opacity=1 ] (388,107.64) .. controls (388,94.73) and (398.59,84.27) .. (411.65,84.27) .. controls (424.71,84.27) and (435.3,94.73) .. (435.3,107.64) .. controls (435.3,120.54) and (424.71,131) .. (411.65,131) .. controls (398.59,131) and (388,120.54) .. (388,107.64) -- cycle ;
\draw  [color={rgb, 255:red, 0; green, 0; blue, 0 }  ,draw opacity=0 ][fill={rgb, 255:red, 23; green, 76; blue, 208 }  ,fill opacity=1 ] (248,105.64) .. controls (248,92.73) and (258.59,82.27) .. (271.65,82.27) .. controls (284.71,82.27) and (295.3,92.73) .. (295.3,105.64) .. controls (295.3,118.54) and (284.71,129) .. (271.65,129) .. controls (258.59,129) and (248,118.54) .. (248,105.64) -- cycle ;
\draw  [color={rgb, 255:red, 0; green, 0; blue, 0 }  ,draw opacity=0 ][fill={rgb, 255:red, 23; green, 76; blue, 208 }  ,fill opacity=1 ] (319,170.64) .. controls (319,157.73) and (329.59,147.27) .. (342.65,147.27) .. controls (355.71,147.27) and (366.3,157.73) .. (366.3,170.64) .. controls (366.3,183.54) and (355.71,194) .. (342.65,194) .. controls (329.59,194) and (319,183.54) .. (319,170.64) -- cycle ;
\draw    (357.9,59.84) -- (394.9,91.84) ;

\draw    (286.9,123.84) -- (323.9,155.84) ;

\draw    (361,155.28) -- (396.05,125.3) ;

\draw    (289,90.28) -- (325.05,60.3) ;

\draw    (435.3,107.64) -- (500.9,107.84) ;

\draw  [color={rgb, 255:red, 0; green, 0; blue, 0 }  ,draw opacity=0 ][fill={rgb, 255:red, 196; green, 11; blue, 18 }  ,fill opacity=1 ] (500.9,107.84) .. controls (500.9,94.94) and (511.49,84.48) .. (524.55,84.48) .. controls (537.61,84.48) and (548.2,94.94) .. (548.2,107.84) .. controls (548.2,120.74) and (537.61,131.2) .. (524.55,131.2) .. controls (511.49,131.2) and (500.9,120.74) .. (500.9,107.84) -- cycle ;
\draw  [color={rgb, 255:red, 0; green, 0; blue, 0 }  ,draw opacity=0 ][fill={rgb, 255:red, 196; green, 11; blue, 18 }  ,fill opacity=1 ] (402.3,236.64) .. controls (402.3,223.73) and (412.89,213.27) .. (425.95,213.27) .. controls (439.01,213.27) and (449.6,223.73) .. (449.6,236.64) .. controls (449.6,249.54) and (439.01,260) .. (425.95,260) .. controls (412.89,260) and (402.3,249.54) .. (402.3,236.64) -- cycle ;
\draw  [color={rgb, 255:red, 0; green, 0; blue, 0 }  ,draw opacity=0 ][fill={rgb, 255:red, 196; green, 11; blue, 18 }  ,fill opacity=1 ] (421.3,29.64) .. controls (421.3,16.73) and (431.89,6.27) .. (444.95,6.27) .. controls (458.01,6.27) and (468.6,16.73) .. (468.6,29.64) .. controls (468.6,42.54) and (458.01,53) .. (444.95,53) .. controls (431.89,53) and (421.3,42.54) .. (421.3,29.64) -- cycle ;
\draw  [color={rgb, 255:red, 0; green, 0; blue, 0 }  ,draw opacity=0 ][fill={rgb, 255:red, 196; green, 11; blue, 18 }  ,fill opacity=1 ] (157.3,61.64) .. controls (157.3,48.73) and (167.89,38.27) .. (180.95,38.27) .. controls (194.01,38.27) and (204.6,48.73) .. (204.6,61.64) .. controls (204.6,74.54) and (194.01,85) .. (180.95,85) .. controls (167.89,85) and (157.3,74.54) .. (157.3,61.64) -- cycle ;
\draw    (202.9,71.84) -- (250.9,96.84) ;

\draw    (362.9,35.84) -- (421.3,29.64) ;

\draw    (360.9,184.84) -- (408.9,219.84) ;

\draw (158,390) node  [align=left] {Target Subgraph};

\end{tikzpicture}
&
{\begin{tikzpicture}[baseline=-23mm,x=0.18pt,y=0.18pt,yscale=-1,xscale=1]

\draw  [color={rgb, 255:red, 0; green, 0; blue, 0 }  ,draw opacity=0 ][fill={rgb, 255:red, 23; green, 76; blue, 208 }  ,fill opacity=1 ] (318,44.64) .. controls (318,31.73) and (328.59,21.27) .. (341.65,21.27) .. controls (354.71,21.27) and (365.3,31.73) .. (365.3,44.64) .. controls (365.3,57.54) and (354.71,68) .. (341.65,68) .. controls (328.59,68) and (318,57.54) .. (318,44.64) -- cycle ;
\draw  [color={rgb, 255:red, 0; green, 0; blue, 0 }  ,draw opacity=0 ][fill={rgb, 255:red, 23; green, 76; blue, 208 }  ,fill opacity=1 ] (392,103.64) .. controls (392,90.73) and (402.59,80.27) .. (415.65,80.27) .. controls (428.71,80.27) and (439.3,90.73) .. (439.3,103.64) .. controls (439.3,116.54) and (428.71,127) .. (415.65,127) .. controls (402.59,127) and (392,116.54) .. (392,103.64) -- cycle ;
\draw  [color={rgb, 255:red, 0; green, 0; blue, 0 }  ,draw opacity=0 ][fill={rgb, 255:red, 23; green, 76; blue, 208 }  ,fill opacity=1 ] (248,105.64) .. controls (248,92.73) and (258.59,82.27) .. (271.65,82.27) .. controls (284.71,82.27) and (295.3,92.73) .. (295.3,105.64) .. controls (295.3,118.54) and (284.71,129) .. (271.65,129) .. controls (258.59,129) and (248,118.54) .. (248,105.64) -- cycle ;
\draw  [color={rgb, 255:red, 0; green, 0; blue, 0 }  ,draw opacity=0 ][fill={rgb, 255:red, 23; green, 76; blue, 208 }  ,fill opacity=1 ] (319,170.64) .. controls (319,157.73) and (329.59,147.27) .. (342.65,147.27) .. controls (355.71,147.27) and (366.3,157.73) .. (366.3,170.64) .. controls (366.3,183.54) and (355.71,194) .. (342.65,194) .. controls (329.59,194) and (319,183.54) .. (319,170.64) -- cycle ;
\draw    (357.9,59.84) -- (394.9,91.84) ;

\draw    (286.9,123.84) -- (323.9,155.84) ;

\draw    (361,155.28) -- (403.9,122.84) ;

\draw    (289,90.28) -- (325.05,60.3) ;

\draw    (426.9,121.84) -- (439.9,184.84) ;

\draw  [color={rgb, 255:red, 0; green, 0; blue, 0 }  ,draw opacity=0 ][fill={rgb, 255:red, 196; green, 11; blue, 18 }  ,fill opacity=1 ] (420.9,204.84) .. controls (420.9,191.94) and (431.49,181.48) .. (444.55,181.48) .. controls (457.61,181.48) and (468.2,191.94) .. (468.2,204.84) .. controls (468.2,217.74) and (457.61,228.2) .. (444.55,228.2) .. controls (431.49,228.2) and (420.9,217.74) .. (420.9,204.84) -- cycle ;
\draw  [color={rgb, 255:red, 0; green, 0; blue, 0 }  ,draw opacity=0 ][fill={rgb, 255:red, 196; green, 11; blue, 18 }  ,fill opacity=1 ] (220.3,233.64) .. controls (220.3,220.73) and (230.89,210.27) .. (243.95,210.27) .. controls (257.01,210.27) and (267.6,220.73) .. (267.6,233.64) .. controls (267.6,246.54) and (257.01,257) .. (243.95,257) .. controls (230.89,257) and (220.3,246.54) .. (220.3,233.64) -- cycle ;
\draw  [color={rgb, 255:red, 0; green, 0; blue, 0 }  ,draw opacity=0 ][fill={rgb, 255:red, 196; green, 11; blue, 18 }  ,fill opacity=1 ] (421.3,29.64) .. controls (421.3,16.73) and (431.89,6.27) .. (444.95,6.27) .. controls (458.01,6.27) and (468.6,16.73) .. (468.6,29.64) .. controls (468.6,42.54) and (458.01,53) .. (444.95,53) .. controls (431.89,53) and (421.3,42.54) .. (421.3,29.64) -- cycle ;
\draw  [color={rgb, 255:red, 0; green, 0; blue, 0 }  ,draw opacity=0 ][fill={rgb, 255:red, 196; green, 11; blue, 18 }  ,fill opacity=1 ] (157.3,61.64) .. controls (157.3,48.73) and (167.89,38.27) .. (180.95,38.27) .. controls (194.01,38.27) and (204.6,48.73) .. (204.6,61.64) .. controls (204.6,74.54) and (194.01,85) .. (180.95,85) .. controls (167.89,85) and (157.3,74.54) .. (157.3,61.64) -- cycle ;
\draw    (202.9,71.84) -- (250.9,96.84) ;

\draw    (362.9,35.84) -- (421.3,29.64) ;

\draw    (322.9,183.84) -- (265.9,225.84) ;

\draw    (341.65,68) -- (342.65,147.27) ;

\draw    (294.1,107.54) -- (391.9,108.84) ;

\end{tikzpicture}}
\end{tabular}

\caption{Left: an example of a subgraph matching problem, where the graph with the blue nodes is matched against the bigger graph. Right: an example of a graph containing a clique. The blue nodes represent a fully connected subgraph of dimension 4, whereas the red nodes do not belong to the clique.}

\label{fig:subgraph}
\end{figure}

Given a graph $G$ and a graph $S$ such that $|S| \le |G|$, the subgraph matching problem consists in finding the nodes of a subgraph $\hat S \subset G$ which is isomorphic to $S$.  
The task is that of learning a function $\tau_S$, such that $\tau_S(G,n)=1, n \in V$, when the node $n$ belongs to the given subgraph $S$, otherwise $\tau_S(G,n)=0$.
The target subgraph $S$ is predefined in the learning phase by providing examples of graphs $G$ that contain $S$ (the nodes of $G$ that define the subgraph $S$ have a supervision equal to 1).
The problem of finding a given subgraph is common in many practical problems and corresponds, for instance, to finding a particular small molecule inside a greater compound.  An example of a subgraph structure is shown in Figure \ref{fig:subgraph} (left).
The dataset the we considered is composed of 100 different graphs, each one having 7 nodes.  The number of nodes of the target subgraph $S$ is instead 3. { The subgraph to be detected is randomly generated in each trial and inserted into every graph of the dataset.}
{ Following \cite{DBLP:journals/tnn/ScarselliGTHM09}, the nodes have integer nodal
features in the range $[0,10$], with the addition of a small normal noise, with zero mean and a standard deviation of 0.25. The LP-GNN model must be capable to exploit both the topological connectivity and the relative distribution of the features in $S$ with respect to the features in the graphs.}

%

\begin{table}[th!]
 \caption{Accuracies on the artificial datasets, for the proposed model (Lagrangian Propagation GNN - LP-GNN) and the standard {GNN*}  model for different settings.}
    \centering

\begin{tabular}{ccc|p{0.8cm}p{0.8cm}|p{0.8cm}p{0.8cm}}
\toprule
\multirow{2}{*}{Model} & & & \multicolumn{2}{c}{Subgraph} & \multicolumn{2}{c}{Clique} \\
\cmidrule{2-7}
& $\mathcal{G}$ & $\epsilon$ &    \textit{Acc(avg)}       &   \textit{Acc(std)}  & \textit{Acc(avg)}       &   \textit{Acc(std)} \\
\midrule

\multirow{7}{*}{LP-GNN} & \multirow{3}{*}{\textit{abs}} & 0.00 &  96.25 &  0.96  &  88.80 &  4.82 \\
 &  & 0.01 &  \textbf{96.30} &  0.87 &  88.75 &  5.03\\
 &  & 0.10 &  95.80 &  0.85 &  85.88 &  4.13 \\
  \cmidrule{2-7}
 & \multirow{3}{*}{\textit{lin}}  & 0.00 &  95.94 & 0.91 &   84.61 &  2.49 \\
 & & 0.01 &  95.94 &  0.91 &  85.21 &  0.54 \\
 & & 0.10 &  95.80 &  0.85 &  85.14 &  2.17 \\
  \cmidrule{2-7}
 & \textit{squared} & - &  96.17 &  1.01 &   \textbf{93.07} &  2.18 \\
 \midrule
{GNN*}  \cite{DBLP:journals/tnn/ScarselliGTHM09} & - & - & 95.86 &  0.64 &  91.86 &  1.12\\
\bottomrule

\end{tabular}

   \label{tab:artificial_results}
\end{table}

\paragraph{\textbf{Clique localization}}
A clique is a complete graph, i.e. a graph in which each node is connected with all the others. In a network, overlapping cliques (i.e. cliques that share some nodes) are admitted. 
Clique localization is a particular instance of the subgraph matching problem, with $S$ being complete. However, the several symmetries contained in a clique makes the graph isomorphism test more difficult. Indeed, it is known that the graph isomorphism has polynomial time solutions only in absence of symmetries.
A clique example is shown in Figure \ref{fig:subgraph} (right).
In the experiments, we consider a dataset composed by graphs having 7 nodes each, where the dimension of the maximal clique is 3 nodes. { The  nodal features are integer random values in the range $[0,8]$, as in \cite{gori2005new}.}

We designed a batch of experiments on these two tasks aimed at validating our simple local optimization approach to constraint-based networks. In particular, we want to show that our optimization scheme can learn better transition and output functions than the corresponding {GNN*} of \cite{DBLP:journals/tnn/ScarselliGTHM09}. Moreover, we want to investigate the behaviour of the algorithm for different choices of the function $\mathcal{G}(x)$, i.e. when changing how we enforce the state convergence constraints. In particular, we tested functions with different properties: \textit{$\epsilon$-insensitive functions}, i.e $\mathcal{G}(x)=0,\  \forall x: -\epsilon \le x \le \epsilon$,  \textit{unilateral functions}, i.e. $\mathcal{G}(x) \in \mathbb{R}^+$, and \textit{bilateral functions}, i.e. $\mathcal{G}(x) \in \mathbb{R}$ (a $\mathcal{G}$ function is either unilateral or bilateral). Table~\ref{tab:g_functions} reports the definition of the considered functions showing if they are $\epsilon$-insensitive, bilateral or unilateral.

Following the experimental setting of \cite{DBLP:journals/tnn/ScarselliGTHM09}, we exploited a training, validation and test set having the same size, i.e. 100 graphs each. We tuned the hyperparameters on the validation data, by selecting the node state dimension from the set  $\{5, 10, 35\}$,  the dropout drop-rate from the set $ \{0, 0.7 \}$,  the state \textit{transition function} from $\{f_{a,v}^{\text{(SUM)}}, f_{a,v}^{\text{(AVG)}}\}$, where
\begin{align*}
 f_{a,v}^{\text{(SUM)}} & = \sum_{u \in ne[v]} h(x_u,l_u,l_{(v,u)},l_{(u,v)},x_v,l_v\ |\ \theta_h) \\
     f_{a,v}^{\text{(AVG)}} & = \tfrac{1}{|ne[v]|}\hskip-2mm \sum_{u \in ne[v]} \hskip-3mm h(x_u,l_u,l_{(v,u)},l_{(u,v)},x_v,l_v\ |\ \theta_h),
\end{align*}
that are based on some of the aggregation functions of Table \ref{tab:gnn}.
Their number of hidden units  was selected from  $ \{5, 20, 50\}$. 
The learning rate for parameters $\theta_{f_a}$ and $\theta_{f_r}$ is selected from the set $\{10^{-5}, 10^{-4},10^{-3}\}$, and the learning rate for the variables $x_v$ and $\lambda_v$ from the set $ \{ 10^{-4}, 10^{-3}, 10^{-2}\}$.

We compared our model with the equivalent {GNN*} in \cite{DBLP:journals/tnn/ScarselliGTHM09}, with the same number of hidden neurons of the $f_a$ and $f_r$ functions. Results are presented in Table~\ref{tab:artificial_results}. On average, LP-GNN perform favourably than vanilla {GNN*}  when the $\mathcal{G}$ function is properly selected.
Constraints characterized by \textit{unilateral functions} usually offer better performances than equivalent bilateral constraints. This might be due to the fact that keeping constraints positive (as in unilateral constraints) provides a more stable learning process. Moreover, smoother constraints (i.e \textit{squared}) or $\epsilon$-insensitive constraints tend to perform slightly better than the other versions. This can be due to the fact that as the constraints move closer to 0 they tend to give a small or null contribution, for \textit{squared} and \textit{abs-$\epsilon$} respectively, acting as regularizers. 


\subsection{Graph Classification}
\label{sec:graph_benchmarks}

\begin{table*}[th!]\

\centering
 \caption{Average and standard deviation of the classification accuracy on the graph classification benchmarks, evaluated on the test set, for different GNN models. The proposed model is denoted as LP-GNN and it is evaluated in two different configurations. LP-GNN-Single exploits only one layer of the diffusion mechanism, while LP-GNN-Multi exploits multiple layers as described in Section \ref{sec:layered}. It is interesting to note that, even if LP-GNN-Single exploits only a shallow representation of nodes, it performs, on average, on-par with respect to other state-of-the-art models. Finally, the LP-GNN-Multi model performs equally to or better than most of the competitors on most of the benchmarks.}
\vspace{1em}
\begin{tabular}{@{}lccccccc@{}}\toprule
 Datasets &  {\textsc{IMDB-B}} & {\textsc{IMDB-M}}  & {\textsc{MUTAG}} &  {\textsc{PROT.}}  & {\textsc{PTC}} & {\textsc{NCI1}}  \\
 \text{\# graphs }  & 1000  & 1500  &  188 & 1113 & 344  &  4110     \\
 \text{\# classes }   &  2  & 3  &   2 & 2  &  2   & 2 \\
 \text{Avg \# nodes }  &  19.8   & 13.0  &  17.9 & 39.1  & 25.5  &  29.8  
\\ \midrule
 \textsc{DCNN}             &  49.1 & 33.5     &   67.0 & 61.3 & 56.6 &  62.6 \\
 \textsc{PatchySan}        & 71.0 $\pm$ 2.2  & 45.2 $\pm$ 2.8  & { \textbf{92.6 }$\pm$ 4.2}  & 75.9 $\pm$ 2.8  & 60.0 $\pm$ 4.8   &  78.6 $\pm$ 1.9     \\ 
 \textsc{DGCNN}           & 70.0 &  47.8     & 85.8        &  75.5   & 58.6  & 74.4   \\ 
 \textsc{AWL}               & 74.5 $\pm$ 5.9  & 51.5 $\pm$ 3.6    & 87.9 $\pm$ 9.8        & -- & --  & --  \\
 \textsc{GraphSAGE}               & 72.3 $\pm$ 5.3  & 50.9 $\pm$ 2.2    & 85.1 $\pm$ 7.6        & 75.9 $\pm$ 3.2 & 63.9 $\pm$ 7.7 & 77.7 $\pm$ 1.5  \\
 \textsc{GIN}   &  { 75.1 $\pm$ 5.1}     & { \textbf{52.3} $\pm$ 2.8}    &  { 89.4 $\pm$ 5.6}     & { 76.2 $\pm$ 2.8}    & { 64.6 $\pm$ 7.0}  & { \textbf{82.7} $\pm$ 1.7}\\
  \textsc{{GNN*}} & 60.9  $\pm$ 5.7  & 41.1  $\pm$ 3.8 & 88.8  $\pm$ 11.5 & 76.4  $\pm$ 4.4 & 61.2  $\pm$ 8.5 & 51.5  $\pm$ 2.6\\
  \textsc{LP-GNN-Single} & 65.3  $\pm$ 4.7 &  46.6  $\pm$ 3.7 & 90.5  $\pm$ 7.0 & 77.1  $\pm$ 4.3  & 64.4  $\pm$ 5.9 & 68.4 $\pm$ 2.1\\
  \textsc{LP-GNN-Multi} & \textbf{76.2}  $\pm$ 3.2 &  51.1  $\pm$ 2.1 & \textbf{92.2}  $\pm$ 5.6 & \textbf{77.5}  $\pm$ 5.2  & \textbf{67.9}  $\pm$ 7.2 & 74.9  $\pm$ 2.4\\

\bottomrule
\end{tabular}

 \label{tab:results}
\end{table*}



Graph-focused tasks consists in finding a common representation of the current input  graph, yielding a single output, as stated by Eq.~(\ref{eq:readout_graph}).
To extract this unique embedding from the representations encoded by all the states available at each node, we implemented the following version of the \textit{readout} function.
\begin{align}
 y_G =  f_r^{\text{(SUM)}} \left(\{x_v, v \in V\}\ |\ \theta_{f_r}\right) & = f_r\left( \sum_{v \in V} f_{a,v} \ |\ \theta_{f_r}\right).
\end{align}
We selected 6 datasets that are popular for benchmarking GNN models.
In particular, four of them are from bioinformatics (MUTAG, PTC, NCI1, PROTEINS) and two from social network analysis (IMDB-BINARY, IMDB-MULTI)~\cite{yanardag2015deep}.

The MUTAG dataset is composed of 188 mutagenic aromatic and heteroaromatic
nitro compounds, having 7 discrete labels. PTC is characterized by
344 chemical compounds belonging to 19 discrete labels, reporting the carcinogenicity for male and female rats. The National Cancer Institute (NCI) made publicly available the NCI1 dataset (4100 nodes), consisting of chemical compounds screened for their ability to
suppress or inhibit the growth of a panel of human tumor cell lines, having 37 discrete labels. 
Nodes in the PROTEINS dataset represent secondary structure elements (SSEs), with an edge connecting them if they are neighbors in the amino-acid
sequence or in the 3D space. The set has 3 discrete labels, representing \textit{helix},
\textit{sheet} or \textit{turn}.
Regarding the social network datasets, IMDB-BINARY is a movie collaboration dataset  collecting actor/actress and genre information of different movies extracted from the popular site IMDB.
For each graph, nodes represent actors/actresses, connected with edges if they appear in the same movie. Each graph is derived from a pre-specified movie category, and the task is to classify the genre it is derived from.  In the IMDB-B dataset, the  collaboration graphs are labelled with the two genres \textit{Action} and \textit{Romance}.
The multi-class version of this dataset is IMDB-MULTI, which is composed by a balanced set of graphs belonging to the \textit{Comedy}, \textit{Romance} and \textit{Sci-Fi} labels.
 
The main purpose of this kind of experiments is to show the ability of the model to strongly exploit the graph topology and structure.  
In fact, whilst in the bioinformatics graphs the nodes have categorical input labels ($l^0_v$) (e.g.  atom  symbol), in the social networks sets there are no input node labels. In this case, we followed what has been recently proposed in \cite{DBLP:journals/corr/abs-1810-00826}, i.e. using one-hot encodings of node degrees. Dataset statistics are summarized in Table \ref{tab:results}.

We compared the proposed Lagrangian Propagation GNN (LP-GNN) scheme with some of the state-of-the-art neural models for graph classification, such as Graph Convolutional Neural Networks. In particular, the models used in the comparison are: 
Diffusion-Convolutional Neural Networks (DCNN)~\cite{DBLP:conf/nips/AtwoodT16}, PATCHY-SAN \cite{DBLP:conf/icml/NiepertAK16}, Deep Graph CNN (DGCNN) \cite{DBLP:conf/aaai/ZhangCNC18}, AWL \cite{ivanov2018anonymous}, GraphSAGE \cite{hamilton2017inductive},  GIN \cite{DBLP:journals/corr/abs-1810-00826}, original {GNN*}  \cite{DBLP:journals/tnn/ScarselliGTHM09}. For all the GNN-like models there are a number of layers equal to 5. We compared also two versions of our scheme: LP-GNN-Single, which is a shallow architecture with $K=1$, and LP-GNN-Multi, which is a layered version of our model, as described in Section \ref{sec:layered}. It is important to notice that differently from LP-GNN-Single, all the convolutional models use a different transition function at each layer. This fact entails that, at a cost of a much larger number of parameters, they have a much higher representational power. Apart from original GNN, we report the accuracy as available in the referred papers.

We followed the evaluation setting of  \cite{DBLP:conf/icml/NiepertAK16}. In particular, we performed 10-fold cross-validation and reported both the average and standard deviation of the accuracies across the 10 folds within the cross-validation. The stopping epoch is selected as the epoch with the best cross-validation accuracy averaged over the 10 folds. 
We tuned the hyperparameters by searching: (1) the number of hidden units for both the $f_a$ and $f_r$ functions from the set $\{5,20,50, 70, 150\}$; (2)  the state transition function from  $\{f_{a,v}^{\text{(SUM)}}, f_{a,v}^{\text{(AVG)}}\}$; (3) the dropout ratio from $\{0, 0.7\}$; (4) the size of the node state $x_v$ from $\{ 10, 35, 50, 70, 150\}$; (5) learning rates for both the $\theta_{f_a}$, $\theta_{f_r}$, $x_v$ and $\lambda_v$ from $\{ 0.1, 0.01, 0.001 \}$. 

Results are reported in Table \ref{tab:results}. As  previously  stated and as it will be further discussed in Section \ref{sec:depth_vs_diffusion}, the LP-GNN-Single model offers performances that, on average, are preferable or on-par to the ones obtained by more complex models that exploit a larger amount of parameters. Finally, the LP-GNN-Multi model performs equally to or better than most of the competitors on most of the benchmarks.

{
\subsection{Non-Isomorphic Graph Classification}
\label{sec:csl}
In Section \ref{sec:com_expr} we discussed the representational capabilities of the proposed model. Even the most expressive aggregation functions available in MPNNs (i.e., GIN, which is exactly as powerful as the 1-WL test) may lack a sufficient  representational power to devise node or edge isomorphism. To address this weakness, in \cite{DBLP:journals/corr/abs-2003-00982} the authors propose to enrich the nodal features with positional encodings capable of describing relative distances and node positioning, in alternative to the usage of more complex GNN models (WL-GNN family \cite{morris2019weisfeiler}). In particular, each attribute is enriched with the $k$ smallest graph Laplacian eigenvectors, in order to obtain smooth encoding description of local neighborhoods. 

In this section, we test if the LP-GNN is capable of leveraging such additional nodal features in the case of the Circulant Skip Links (CSL) 
\cite{murphy2019relational}, a synthetic dataset composed by automorphic graphs. Each graph is a 4-regular cycle graph with skip links, and the dataset is composed by $150$ graphs, balanced in order to contain $15$ graphs for every isomorphism class. Indeed, graphs belonging to the same  isomorphic class share the same \textit{skip length}. Hence, correctly classifying their skip length foster the classification of non-isomorphic graphs.  We considered the exact same splits of  \cite{DBLP:journals/corr/abs-2003-00982}, obtained by a $5$-fold cross validation with stratified sampling.
For each of the $5$ training sets we considered 20 different initializations of the model parameters, reporting the average and the standard deviation of the results. We tuned the hyper-parameters considering the following grid of values: the number of hidden units for both the $f_a$ and $f_r$ functions in $\{20, 40, 80, 100\}$; the $\mathcal{G}$ function in \{$abs-\epsilon, squared\}$  ; the size of the node state $x_v$ in $\{ 25, 50, 100\}$; learning rates for both the $\theta_{f_a}$, $\theta_{f_r}$, $x_v$ and $\lambda_v$ in $\{ 0.1, 0.05, 0.00001 \}$; the LP-GNN and state transition functions number of layers in $\{ 1, 2, 5\}$. We considered 20 thousands epochs with an early stopping given by a patience on the validation loss of 5 thousands epochs. 
Table \ref{tab:csl} shows how LP-GNN are capable of exploiting the rich nodal features constituted by the positional encoding with Laplacian eigenmaps, allowing the model to correctly classify  the graph skip length, hence the isomorphism class. 
\begin{table}[th!]
\centering
 \caption{Average and standard deviation of the classification accuracy on the CSL dataset with positional embeddings, evaluated on the test sets and, for completeness, on the training sets, for different GNNs (results of the competitors are taken from \cite{DBLP:journals/corr/abs-2003-00982}).}
\vspace{1em}

\begin{tabular}{@{}llHHlHH@{}}\toprule 
\multirow{1}{*}{Model} & \multicolumn{3}{l}{Test Accuracy} & \multicolumn{3}{l} {Train Accuracy} \\  
 \midrule
 MLP & $22.57 \pm 6.09 $ & $46.67$ & $10.00$ &  $30.39 \pm 5.71$ & 43.33 & 10.00 \\
 GCN & $100.00 \pm 0.00 $ & $100.00$ & $100.00$ &  $100.00 \pm 0.00$ & 100.003 & 0.00 \\
 GraphSage & $99.93 \pm 0.47 $ & $100.00$ & $96.67$ &  $100.00 \pm 0.00$ & 100.00 & 100.00 \\
 GAT & $99.93 \pm 0.47 $ & $100.00$ & $96.67$ &  $100.00\pm 0.00$ & 100.00 & 100.00 \\
GIN & $99.33 \pm 1.33 $ & $100.00$ & $ 96.67$ &  $100.00 \pm 0.00$ & 100.00 & 100.00 \\
 RingGNN & $17.23 \pm 6.36 $ & $40.00$ & $10.00$ &  $26.12 \pm 14.38$ &  58.89 & 10.00 \\
 3WLGNN & $30.53 \pm 9.86 $ & $56.67$ & $10.00$ &  $99.64 \pm 1.68$ & 100.00 & 88.89 \\
 LP-GNN & $98.90 \pm 4.49 $ & $100.00$ & $56.67$ &  $99.92 \pm 0.67$ & 100.00 & 93.33 \\

\bottomrule
\end{tabular}

 \label{tab:csl}
\end{table}

}

{
\subsection{Standardized Graph Benchmark}
\label{sec:enzymes}

Several works in recent literature \cite{shchur2018pitfalls,errica2020fair,DBLP:journals/corr/abs-2003-00982}  highlighted the lack of a common evaluation protocol for measuring the performances of GNN models. 
In order test the performances of our model in a setting in which the statistical relevance of the dataset is assured, we exploited the data splits provided by the common infrastructure of \cite{DBLP:journals/corr/abs-2003-00982}, focussing on ENZYMES, one of the conventional datasets for benchmarking GNNs. This dataset consist in a set of proteins, represented as graphs, where each node correspond to secondary structure elements. The node connectivity follows the same rules as the previously discussed PROTEINS dataset, whilst each node, in addition to a discrete attribute indicating the structure it belongs (helix, sheet or turn), is enriched with chemical and physical measurements resulting in vector-valued features of size $18$ (secondary structure length, hydrophobicity, polarity, polarizability, etc.). The goal of the task is to assign the input graphs  to  one of the $6$ EC top-level classes, which reflect the catalyzed chemical reaction. 

We considered the same splits of \cite{DBLP:journals/corr/abs-2003-00982}, that are based on 10-fold cross validation, stratified sampling. Each training set is composed by 480 graphs, while the validation and test sets are composed by 60 graphs each.
We tuned the hyper-parameters in the following grids: the number of hidden units for both the $f_a$ and $f_r$ functions in set $\{100, 120, 150, 180\}$; (the $\mathcal{G}$ function in  \{$abs-\epsilon, squared\}$; the size of the node state $x_v$ in $\{ 50, 100, 200\}$; learning rates for both the $\theta_{f_a}$, $\theta_{f_r}$, $x_v$ and $\lambda_v$ in $\{ 0.01, 0.02, 0.0001, 0.0002 \}$; the LP-GNN and state transition functions number of layers in $\{ 1, 2, 5\}$. We considered 20 thousands epochs with an early stopping criterion given by a patience on the validation loss of 5 thousands epochs. Experiments are performed considering two distinct seeds to initialize the random number generators.

\begin{table}[th!]

\centering
 \caption{Average and standard deviation of the classification accuracy on the ENZYMES benchmark, evaluated on the test set and, for completeness, on the training set, for different GNN models (results of the competitors are taken from \cite{DBLP:journals/corr/abs-2003-00982}).}
\vspace{1em}

\begin{tabular}{@{}p{1.1cm}p{1.45cm}p{1.6cm}p{1.45cm}p{1.6cm}@{}}\toprule 
\multirow{2}{*}{Model} & \multicolumn{2}{c}{Seed 1} & \multicolumn{2}{c} {Seed 2} \\  
\cmidrule(l){2-3}   \cmidrule(l){4-5}   
 &  Test Acc.  & Train Acc.   &  Test Acc.  & Train Acc.  \\
 \midrule
 MLP & $55.83 \pm 3.52 $ & $93.06 \pm 7.55$ & $53.83 \pm 4.72$ &  $87.85 \pm  10.76$ \\ 
  GCN & $65.83 \pm 4.61 $ & $97.69 \pm 3.06$ & $64.83 \pm 7.09$ &  $93.04 \pm  4.98$ \\ 
   GraphSage & $65.00 \pm 4.94 $ & $100.00 \pm 0.00$ & $68.17 \pm 5.45$ &  $100.00 \pm  0.00$ \\ 
    GAT & $68.50 \pm 5.24 $ & $100.00 \pm 0.00$ & $68.50 \pm 4.62$ &  $100.00 \pm  0.00$ \\ 
     GIN & $65.33 \pm 6.82 $ & $100.00 \pm 0.00$ & $67.66 \pm 5.83$ &  $100.00 \pm  0.00$ \\ 
      RingGNN & $18.67 \pm 1.79 $ & $ 20.10 \pm 2.17$ & $45.33 \pm 4.52$ &  $ 56.79 \pm  6.08$ \\ 
     3WLGNN & $61.00 \pm 6.80 $ & $98.87 \pm 1.57$ & $57.67 \pm 9.52$ &  $96.73 \pm  5.52$ \\ 
     LP-GNN & $66.00 \pm 6.09 $ & $93.29 \pm 4.86$ & $64.17 \pm 5.6$ &  $92.52\pm  2.41 $ \\


\bottomrule
\end{tabular}

 \label{tab:enzymes}
\end{table}
{
This experiment confirmed the observations from \cite{DBLP:journals/corr/abs-2003-00982} regarding similar statistical test performance among the various GNN models. Indeed, the proposed LP-GNN  achieves comparable results with respect to the most promising GNN models in both the runs.  
}
}

\subsection{Depth vs Diffusion}
\label{sec:depth_vs_diffusion}

It is interesting to note that for current {ConvGNNs}  models \cite{DBLP:conf/nips/AtwoodT16,DBLP:conf/icml/NiepertAK16,DBLP:conf/aaai/ZhangCNC18,ivanov2018anonymous,DBLP:journals/corr/abs-1810-00826,hamilton2017inductive} the role of the architecture depth is twofold. First, as it is common in deep learning, depth is used to perform a multi-layer feature extraction of node inputs, providing more and more representational power as depth increases. Secondly, it allows node information to flow through the graph fostering the realisation of a diffusion mechanism.
Conversely, our model strictly splits these two processes. Diffusion is realised by looking for a fixed point of the state transition function, while deep feature extraction is realised by stacking multiple layers of  node states, enabling a separate diffusion process at each layer. We believe this distinction to be a fundamental ingredient for a clearer understanding of which mechanism, between diffusion and node deep representation, is concurring in achieving specific performances. 
In the previous section, we showed indeed that our diffusion mechanism paired only with a simple shallow representation of nodes (reffered as LP-GNN-Single) is sufficient, in most cases, to match performances of much deeper and complex networks.
In this section, we want to investigate further this aspect. In particular, we focused on the IMDB-B dataset. The choice of this dataset has to be attributed to the fact that it contains no node features. In this way, as done in Section \ref{sec:qualitative}, we can assure that the only information available to solve the task is the topological one. Other datasets, for example the simpler MUTAG, reach high level accuracies even without the structural information of the graph, by only exploiting node features.

We compared our model with the state-of-the-art GIN \cite{DBLP:journals/corr/abs-1810-00826} model. For both models, we tested four architectures with \{1,2,3,5\} GNN state layers. We want to show that in very shallow GNNs (one layer) our model can still perform fairly well, since the diffusion process is independent from the depth of the network. On the other side, the GIN model, as other graph convolutional networks, needs deep architectures with a larger number of parameters for the diffusion process to take place. We believe this to be a big advantage of our model w.r.t. convolutional architectures in all the cases where high representational power is not required.
Results are shown in Table \ref{tab:depth_vs_diffusion}. It can be noted that this task can reach the 96\% of the top accuracies (75.1 and 76.2, respectively) using only 2 layers of GNN, for both the competitors. The great difference between the two approaches becomes clear in the architecture composed by only 1 layer. In this setting, the GIN model, like all the other convolutional architectures, can only exploit information contained in direct 1-hop neighbors, reaching a 52\% accuracy (which is close to random in a binary classification task). On the contrary, our model can reach a 65.3\% of accuracy (85\% of the maximum accuracy). This is a signal of the fact that convolutional architectures need a second layer (and thus a larger number of parameters) mainly to perform diffusion at 2-hop neighbors rather than for higher representational power.

\begin{table}[]
    \centering
    \begin{tabular}{c|c|cccc}
        \toprule
         \multirow{2}{*}{\textbf{Evaluation}}& \multirow{2}{*}{\textbf{Model}} & \multicolumn{4}{c}{\textbf{Number of State Layers}} \\
         & & 1 & 2 & 3 & 5 \\
         \midrule
         \multirow{2}{*}{Absolute } & GIN \cite{DBLP:journals/corr/abs-1810-00826} & 52 &	72.6 &	72.7 &	75.1\\
         & LP-GNN & \textbf{65.3} &	\textbf{73.7} &	\textbf{73.9}&	\textbf{76.2}\\
         \midrule
         \multirow{2}{*}{Relative }& GIN \cite{DBLP:journals/corr/abs-1810-00826} & 69 &	96 & 97 &	100\\
         & LP-GNN & \textbf{85} &	96 &	97& 100\\
         \bottomrule
    \end{tabular}
    \caption{Absolute and Relative test accuracies in the IMDB-B data when the number of GNN layers varies from 1 to 5 (i.e. $K \in [1,5]$). Relative accuracy is the $\%$ of the current accuracy with respect to the maximum obtained performances.  The state-of-the-art GIN \cite{DBLP:journals/corr/abs-1810-00826} model and our proposed approach are compared.}
    \label{tab:depth_vs_diffusion}
\end{table}

{\subsection{Lagrangian Propagation in  GNNs}
\label{sec:extending_agg_func}}
 { The proposed LP-GNN experimented in this paper exploits the aggregation function of classic GNN* (first row of Table \ref{tab:gnn}). However, the generality of the proposed approach enables, in principle, its extension to a wide variety of aggregation functions available in literature (see the references in Table \ref{tab:gnn}). We adapted the proposed constraint-based scheme to use the aggregation functions of two popular models, i.e., GIN and GCN -- see Table \ref{tab:gnn}. In so doing, we obtained two new models denoted with LP-GIN and LP-GCN, respectively. We tested the performances of such models in the same experimental setting of Section~\ref{sec:depth_vs_diffusion}, and the results are reported in Table~\ref{tab:lpingnns}. Both the new models are capable of leveraging the constraint-based diffusion mechanism, even with a shallow 1-layered architecture, and, vice-versa, the LP-GNN schema proposed in this paper can benefit from the particular aggregation mechanisms of GIN and GCN. 
 

 }

\begin{table}[]
    \centering
    
    \begin{tabular}{c|c|cccc}
        \toprule
         \multirow{2}{*}{\textbf{Evaluation}}& \multirow{2}{*}{\textbf{Model}} & \multicolumn{4}{c}{\textbf{Number of State Layers}} \\
         & & 1 & 2 & 3 & 5 \\
         \midrule
        \multirow{3}{*}{Absolute } & LP-GNN & 65.3 &	73.7 &	73.9&	76.2\\
          & LP-GIN  & 71.3 &	73.2 &	73.0 &	73.6\\
          & LP-GCN  & 73.4 &	73.5 &	73.8 &	73.7\\         
         \bottomrule
    \end{tabular}
    \caption{{  {LP diffusion in other GNN models}. Setting of Table~\ref{tab:depth_vs_diffusion}.}}
    \label{tab:lpingnns}
\end{table}

\vskip 3mm
\section{Conclusions and Future Work}
\label{sec:concl}
\vskip 1mm
We proposed an approach that simplifies the learning procedure of  {the GNN* model of \cite{DBLP:journals/tnn/ScarselliGTHM09}}, making it more easily implementable and more efficient.
The formulation of the learning task as a constrained optimization problem allows us to avoid the explicit computation of the fixed point, that is needed to encode the graph. The proposed framework defines how to jointly optimize the model weights and the state representations without the need {of the separate optimization stages that are typical of GNN*}. 
The constrained learning procedure can be replicated recursively multiple times, thus introducing different levels of abstraction and different representations, similarly to what happens in multi-layer networks.
We proposed and investigated constraining functions that allow the model to modulate the effects of the diffusion process. Our experiments have shown that the proposed approach leads to results that compare favourably with the ones of related models.
{ The proposed constraint-based scheme can be easily exploited in other GNN models, leveraging their aggregation capabilities while diffusing information, as we briefly experimented. 
Future work will include a deeper analysis of the proposed scheme when more specifically customized to the case of other GNN architectures.}





\ifCLASSOPTIONcompsoc
 \section*{Acknowledgments}
This work was partly supported by the PRIN 2017 project RexLearn, funded by the Italian Ministry of Education, University and Research (grant no. 2017TWNMH2). This work was supported by the Research Foundation - Flanders and the KU Leuven Research Fund.


\ifCLASSOPTIONcaptionsoff
  \newpage
\fi



\bibliographystyle{IEEEtran}
\bibliography{refe.bib}

\begin{thebibliography}{10}
\providecommand{\url}[1]{#1}
\csname url@samestyle\endcsname
\providecommand{\newblock}{\relax}
\providecommand{\bibinfo}[2]{#2}
\providecommand{\BIBentrySTDinterwordspacing}{\spaceskip=0pt\relax}
\providecommand{\BIBentryALTinterwordstretchfactor}{4}
\providecommand{\BIBentryALTinterwordspacing}{\spaceskip=\fontdimen2\font plus
\BIBentryALTinterwordstretchfactor\fontdimen3\font minus
  \fontdimen4\font\relax}
\providecommand{\BIBforeignlanguage}[2]{{%
\expandafter\ifx\csname l@#1\endcsname\relax
\typeout{** WARNING: IEEEtran.bst: No hyphenation pattern has been}%
\typeout{** loaded for the language `#1'. Using the pattern for}%
\typeout{** the default language instead.}%
\else
\language=\csname l@#1\endcsname
\fi
#2}}
\providecommand{\BIBdecl}{\relax}
\BIBdecl

\bibitem{Williams1989ALA}
R.~J. Williams and D.~Zipser, ``A learning algorithm for continually running
  fully recurrent neural networks,'' \emph{Neural Computation}, vol.~1, pp.
  270--280, 1989.

\bibitem{hochreiter1997long}
S.~Hochreiter and J.~Schmidhuber, ``Long short-term memory,'' \emph{Neural
  Computation}, vol.~9, no.~8, pp. 1735--1780, 1997.

\bibitem{LeCun:1998:CNI:303568.303704}
Y.~LeCun and Y.~Bengio, ``Convolutional networks for images, speech, and
  time-series,'' in \emph{The Handbook of Brain Theory and Neural Networks},
  M.~A. Arbib, Ed.\hskip 1em plus 0.5em minus 0.4em\relax MIT Press, 1995.

\bibitem{Goller96}
C.~{Goller} and A.~{Kuchler}, ``Learning task-dependent distributed
  representations by backpropagation through structure,'' in
  \emph{International Conference on Neural Networks}, vol.~1, June 1996, pp.
  347--352 vol.1.

\bibitem{Frasconi:1998:GFA:2325763.2326281}
P.~Frasconi, M.~Gori, and A.~Sperduti, ``A general framework for adaptive
  processing of data structures,'' \emph{IEEE Transactions on Neural Networks},
  vol.~9, no.~5, pp. 768--786, 1998.

\bibitem{DBLP:journals/tnn/ScarselliGTHM09}
F.~Scarselli, M.~Gori, A.~C. Tsoi, M.~Hagenbuchner, and G.~Monfardini, ``The
  graph neural network model,'' \emph{{IEEE} Transactions on Neural Networks},
  vol.~20, no.~1, pp. 61--80, 2009.

\bibitem{carreira2014distributed}
M.~Carreira-Perpinan and W.~Wang, ``Distributed optimization of deeply nested
  systems,'' in \emph{Artificial Intelligence and Statistics}, 2014, pp.
  10--19.

\bibitem{taylor2016training}
G.~Taylor, R.~Burmeister, Z.~Xu, B.~Singh, A.~Patel, and T.~Goldstein,
  ``Training neural networks without gradients: A scalable admm approach,'' in
  \emph{International Conference on Machine Learning}, 2016, pp. 2722--2731.

\bibitem{DBLP:journals/tnn/WuPCLZY21}
Z.~Wu, S.~Pan, F.~Chen, G.~Long, C.~Zhang, and P.~S. Yu, ``A comprehensive
  survey on graph neural networks,'' \emph{{IEEE} Transactions on Neural
  Networks and Learning Systems}, vol.~32, no.~1, pp. 4--24, 2021.

\bibitem{Platt}
J.~C. Platt and A.~H. Barr, ``Constrained differential optimization,'' in
  \emph{Neural Information Processing Systems}, 1988, pp. 612--621.

\bibitem{DBLP:conf/ecai/TiezziMMMG20}
M.~Tiezzi, G.~Marra, S.~Melacci, M.~Maggini, and M.~Gori, ``A lagrangian
  approach to information propagation in graph neural networks,'' in \emph{ECAI
  2020 - 24th European Conference on Artificial Intelligence}, ser. Frontiers
  in Artificial Intelligence and Applications, vol. 325.\hskip 1em plus 0.5em
  minus 0.4em\relax IOS Press, 2020, pp. 1539--1546.

\bibitem{DBLP:journals/spm/BronsteinBLSV17}
M.~M. Bronstein, J.~Bruna, Y.~LeCun, A.~Szlam, and P.~Vandergheynst,
  ``Geometric deep learning: Going beyond euclidean data,'' \emph{{IEEE} Signal
  Processing Magazine}, vol.~34, no.~4, pp. 18--42, 2017.

\bibitem{melacci2011primallapsvm}
S.~Melacci and M.~Belkin, ``{Laplacian Support Vector Machines Trained in the
  Primal},'' \emph{Journal of Machine Learning Research}, vol.~12, pp.
  1149--1184, March 2011.

\bibitem{Sperduti:1997:SNN:2325755.2326105}
A.~Sperduti and A.~Starita, ``Supervised neural networks for the classification
  of structures,'' \emph{IEEE Transactions on Neural Networks}, vol.~8, no.~3,
  pp. 714--735, 1997.

\bibitem{DBLP:journals/corr/LiTBZ15}
Y.~Li, D.~Tarlow, M.~Brockschmidt, and R.~S. Zemel, ``Gated graph sequence
  neural networks,'' in \emph{International Conference on Learning
  Representations}, 2016.

\bibitem{DBLP:conf/icml/DaiKDSS18}
H.~Dai, Z.~Kozareva, B.~Dai, A.~J. Smola, and L.~Song, ``Learning steady-states
  of iterative algorithms over graphs,'' in \emph{International Conference on
  Machine Learning}, vol.~80, 2018, pp. 1114--1122.

\bibitem{DBLP:journals/nn/BacciuEMP20}
D.~Bacciu, F.~Errica, A.~Micheli, and M.~Podda, ``A gentle introduction to deep
  learning for graphs,'' \emph{Neural Networks}, vol. 129, pp. 203--221, 2020.

\bibitem{DBLP:conf/iclr/Loukas20}
A.~Loukas, ``What graph neural networks cannot learn: depth vs width,'' in
  \emph{International Conference on Learning Representations}, 2020.

\bibitem{DBLP:conf/aaai/LiHW18}
Q.~Li, Z.~Han, and X.~Wu, ``Deeper insights into graph convolutional networks
  for semi-supervised learning,'' in \emph{{AAAI} Conference on Artificial
  Intelligence}, 2018, pp. 3538--3545.

\bibitem{DBLP:journals/corr/abs-2003-00982}
\BIBentryALTinterwordspacing
V.~P. Dwivedi, C.~K. Joshi, T.~Laurent, Y.~Bengio, and X.~Bresson,
  ``Benchmarking graph neural networks,'' \emph{CoRR}, vol. abs/2003.00982,
  2020. [Online]. Available: \url{https://arxiv.org/abs/2003.00982}
\BIBentrySTDinterwordspacing

\bibitem{DBLP:journals/corr/BrunaZSL13}
J.~Bruna, W.~Zaremba, A.~Szlam, and Y.~LeCun, ``Spectral networks and locally
  connected networks on graphs,'' in \emph{International Conference on Learning
  Representations}, 2014.

\bibitem{DBLP:journals/corr/HenaffBL15}
\BIBentryALTinterwordspacing
M.~Henaff, J.~Bruna, and Y.~LeCun, ``Deep convolutional networks on
  graph-structured data,'' \emph{CoRR}, vol. abs/1506.05163, 2015. [Online].
  Available: \url{http://arxiv.org/abs/1506.05163}
\BIBentrySTDinterwordspacing

\bibitem{DBLP:conf/nips/DefferrardBV16}
M.~Defferrard, X.~Bresson, and P.~Vandergheynst, ``Convolutional neural
  networks on graphs with fast localized spectral filtering,'' in
  \emph{Advances in Neural Information Processing Systems}, 2016, pp.
  3837--3845.

\bibitem{DBLP:conf/iclr/KipfW17}
T.~N. Kipf and M.~Welling, ``Semi-supervised classification with graph
  convolutional networks,'' in \emph{International Conference on Learning
  Representations}, 2017.

\bibitem{DBLP:conf/nips/AtwoodT16}
J.~Atwood and D.~Towsley, ``Diffusion-convolutional neural networks,'' in
  \emph{Advances in Neural Information Processing Systems}, 2016, pp.
  1993--2001.

\bibitem{hamilton2017inductive}
W.~L. Hamilton, R.~Ying, and J.~Leskovec, ``Inductive representation learning
  on large graphs,'' in \emph{Advances in Neural Information Processing
  Systems}, 2017.

\bibitem{DBLP:conf/icml/NiepertAK16}
M.~Niepert, M.~Ahmed, and K.~Kutzkov, ``Learning convolutional neural networks
  for graphs,'' in \emph{International Conference on Machine Learning}, 2016,
  pp. 2014--2023.

\bibitem{DBLP:conf/nips/DuvenaudMABHAA15}
D.~K. Duvenaud, D.~Maclaurin, J.~Aguilera{-}Iparraguirre,
  R.~G{\'{o}}mez{-}Bombarelli, T.~Hirzel, A.~Aspuru{-}Guzik, and R.~P. Adams,
  ``Convolutional networks on graphs for learning molecular fingerprints,'' in
  \emph{Advances in Neural Information Processing Systems}, 2015, pp.
  2224--2232.

\bibitem{DBLP:conf/aaai/ZhangCNC18}
M.~Zhang, Z.~Cui, M.~Neumann, and Y.~Chen, ``An end-to-end deep learning
  architecture for graph classification,'' in \emph{{AAAI} Conference on
  Artificial Intelligence}, 2018, pp. 4438--4445.

\bibitem{ivanov2018anonymous}
S.~Ivanov and E.~Burnaev, ``Anonymous walk embeddings,'' in \emph{International
  Conference on Machine Learning}, 2018, pp. 2186--2195.

\bibitem{DBLP:journals/corr/abs-2005-03675}
\BIBentryALTinterwordspacing
I.~Chami, S.~Abu{-}El{-}Haija, B.~Perozzi, C.~R{\'{e}}, and K.~Murphy,
  ``Machine learning on graphs: {A} model and comprehensive taxonomy,''
  \emph{CoRR}, vol. abs/2005.03675, 2020. [Online]. Available:
  \url{https://arxiv.org/abs/2005.03675}
\BIBentrySTDinterwordspacing

\bibitem{DBLP:journals/tsp/GamaMLR19}
F.~Gama, A.~G. Marques, G.~Leus, and A.~Ribeiro, ``Convolutional neural network
  architectures for signals supported on graphs,'' \emph{{IEEE} Transactions on
  Signal Processing}, vol.~67, no.~4, pp. 1034--1049, 2019.

\bibitem{DBLP:journals/tsp/RuizGMR20}
L.~Ruiz, F.~Gama, A.~G. Marques, and A.~Ribeiro, ``Invariance-preserving
  localized activation functions for graph neural networks,'' \emph{{IEEE}
  Transactions on Signal Processing}, vol.~68, pp. 127--141, 2020.

\bibitem{gilmer2017neural}
J.~Gilmer, S.~S. Schoenholz, P.~F. Riley, O.~Vinyals, and G.~E. Dahl, ``Neural
  message passing for quantum chemistry,'' in \emph{International Conference on
  Machine Learning}.\hskip 1em plus 0.5em minus 0.4em\relax PMLR, 2017, pp.
  1263--1272.

\bibitem{gilmer2020message}
J.~Gilmer, S.~Schoenholz, P.~Riley, O.~Vinyals, and D.~George, ``Message
  passing neural networks,'' in \emph{Machine Learning Meets Quantum
  Physics}.\hskip 1em plus 0.5em minus 0.4em\relax Springer, 2020, pp.
  199--214.

\bibitem{velivckovic2018graph}
\BIBentryALTinterwordspacing
P.~Veličković, G.~Cucurull, A.~Casanova, A.~Romero, P.~Liò, and Y.~Bengio,
  ``Graph attention networks,'' in \emph{International Conference on Learning
  Representations}, 2018. [Online]. Available:
  \url{https://openreview.net/forum?id=rJXMpikCZ}
\BIBentrySTDinterwordspacing

\bibitem{monti2017geometric}
F.~Monti, D.~Boscaini, J.~Masci, E.~Rodola, J.~Svoboda, and M.~M. Bronstein,
  ``Geometric deep learning on graphs and manifolds using mixture model cnns,''
  in \emph{IEEE Conference on Computer Vision and Pattern Recognition}, 2017,
  pp. 5115--5124.

\bibitem{bresson2017residual}
\BIBentryALTinterwordspacing
X.~Bresson and T.~Laurent, ``Residual gated graph convnets,'' \emph{CoRR}, vol.
  abs/1711.07553, 2017. [Online]. Available:
  \url{http://arxiv.org/abs/1711.07553}
\BIBentrySTDinterwordspacing

\bibitem{DBLP:journals/corr/abs-1810-00826}
K.~Xu, W.~Hu, J.~Leskovec, and S.~Jegelka, ``How powerful are graph neural
  networks?'' in \emph{International Conference on Learning Representations},
  2019.

\bibitem{weisfeiler1968reduction}
B.~Weisfeiler and A.~A. Lehman, ``A reduction of a graph to a canonical form
  and an algebra arising during this reduction,'' \emph{Nauchno-Technicheskaya
  Informatsia}, vol.~2, no.~9, pp. 12--16, 1968.

\bibitem{morris2019weisfeiler}
C.~Morris, M.~Ritzert, M.~Fey, W.~L. Hamilton, J.~E. Lenssen, G.~Rattan, and
  M.~Grohe, ``Weisfeiler and leman go neural: Higher-order graph neural
  networks,'' in \emph{AAAI Conference on Artificial Intelligence}, vol.~33,
  2019, pp. 4602--4609.

\bibitem{maron2018invariant}
H.~Maron, H.~Ben{-}Hamu, N.~Shamir, and Y.~Lipman, ``Invariant and equivariant
  graph networks,'' in \emph{International Conference on Learning
  Representations}, 2019.

\bibitem{corso2020principal}
G.~Corso, L.~Cavalleri, D.~Beaini, P.~Li{\`{o}}, and P.~Velickovic, ``Principal
  neighbourhood aggregation for graph nets,'' in \emph{Advances in Neural
  Information Processing Systems}, 2020.

\bibitem{lecun1988theoretical}
Y.~LeCun, D.~Touresky, G.~Hinton, and T.~Sejnowski, ``A theoretical framework
  for back-propagation,'' in \emph{Connectionist Models Summer School}, vol.~1,
  CMU, Pittsburgh.\hskip 1em plus 0.5em minus 0.4em\relax Morgan Kaufmann,
  1988, pp. 21--28.

\bibitem{noia}
A.~Gotmare, V.~Thomas, J.~Brea, and M.~Jaggi, ``Decoupling backpropagation
  using constrained optimization methods,'' in \emph{Workshop on Efficient
  Credit Assignment in Deep Learning and Deep Reinforcement Learning,
  International Conference on Machine Learning}, 2018, pp. 1--11.

\bibitem{marra2020local}
G.~Marra, M.~Tiezzi, S.~Melacci, A.~Betti, M.~Maggini, and M.~Gori, ``Local
  propagation in constraint-based neural networks,'' in \emph{International
  Joint Conference on Neural Networks}.\hskip 1em plus 0.5em minus 0.4em\relax
  IEEE, 2020, pp. 1--8.

\bibitem{Bianchini2018}
M.~Bianchini, G.~M. Dimitri, M.~Maggini, and F.~Scarselli, \emph{Deep Neural
  Networks for Structured Data}.\hskip 1em plus 0.5em minus 0.4em\relax
  Springer International Publishing, 2018, pp. 29--51.

\bibitem{kingma2014adam}
D.~P. Kingma and J.~Ba, ``Adam: {A} method for stochastic optimization,'' in
  \emph{International Conference on Learning Representations}, 2015.

\bibitem{rossi2018inductive}
A.~Rossi, M.~Tiezzi, G.~M. Dimitri, M.~Bianchini, M.~Maggini, and F.~Scarselli,
  ``Inductive--transductive learning with graph neural networks,'' in
  \emph{IAPR Workshop on Artificial Neural Networks in Pattern
  Recognition}.\hskip 1em plus 0.5em minus 0.4em\relax Springer, 2018, pp.
  201--212.

\bibitem{zachary1977information}
W.~W. Zachary, ``An information flow model for conflict and fission in small
  groups,'' \emph{Journal of Anthropological Research}, vol.~33, no.~4, pp.
  452--473, 1977.

\bibitem{gori2005new}
M.~Gori, G.~Monfardini, and F.~Scarselli, ``A new model for learning in graph
  domains,'' in \emph{International Joint Conference on Neural Networks},
  vol.~2.\hskip 1em plus 0.5em minus 0.4em\relax IEEE, 2005, pp. 729--734.

\bibitem{yanardag2015deep}
P.~Yanardag and S.~Vishwanathan, ``Deep graph kernels,'' in \emph{International
  Conference on Knowledge Discovery and Data Mining}, 2015, pp. 1365--1374.

\bibitem{murphy2019relational}
R.~Murphy, B.~Srinivasan, V.~Rao, and B.~Ribeiro, ``Relational pooling for
  graph representations,'' in \emph{International Conference on Machine
  Learning}, 2019, pp. 4663--4673.

\bibitem{shchur2018pitfalls}
\BIBentryALTinterwordspacing
O.~Shchur, M.~Mumme, A.~Bojchevski, and S.~G{\"{u}}nnemann, ``Pitfalls of graph
  neural network evaluation,'' \emph{CoRR}, vol. abs/1811.05868, 2018.
  [Online]. Available: \url{http://arxiv.org/abs/1811.05868}
\BIBentrySTDinterwordspacing

\bibitem{errica2020fair}
F.~Errica, M.~Podda, D.~Bacciu, and A.~Micheli, ``A fair comparison of graph
  neural networks for graph classification,'' in \emph{International Conference
  on Learning Representations}, 2020.

\end{thebibliography}
%
%
%

%


\end{document}